\PassOptionsToPackage{dvipsnames,table}{xcolor}
\documentclass{article}

\usepackage{microtype}
\usepackage{graphicx}
\usepackage{subcaption}
\usepackage{booktabs} 

\usepackage{hyperref}

\usepackage[accepted]{icml2026}

\usepackage{amsmath}
\usepackage{amssymb}
\usepackage{mathtools}
\usepackage{amsthm}

\usepackage{xcolor}   
\usepackage{soul}     

\definecolor{mydarkblue}{rgb}{0,0.08,0.45}
\definecolor{darkgreen}{rgb}{0.0, 0.5, 0.0} 
\definecolor{myblue}{RGB}{235,235,250}
\definecolor{lightpink}{RGB}{222, 235, 225} 
\definecolor{lightblue}{RGB}{230, 235, 245} 
\definecolor{lightgray}{RGB}{240, 240, 240} 
\definecolor{darkgray}{RGB}{220, 220, 220} 

\definecolor{superlightred}{rgb}{0.99, 0.92, 0.92}
\definecolor{darkgreen}{RGB}{50,100,0}
\definecolor{darkred}{RGB}{200, 0, 0}

\newcommand{\up}[1]{\textcolor{OliveGreen}{\footnotesize \ $\uparrow${#1}}}

\usepackage[capitalize,noabbrev]{cleveref}

\theoremstyle{plain}

\theoremstyle{definition}

\theoremstyle{remark}

\usepackage[textsize=tiny]{todonotes}

\usepackage{graphicx}   
\usepackage{multirow}   
\usepackage{colortbl}   
\usepackage{booktabs}   
\usepackage{array}      

\usepackage{caption}    

\usepackage{amsmath}     
\usepackage{caption}    
\usepackage{adjustbox}
\usepackage{multirow}
\usepackage{multicol}
\usepackage{subcaption}
\usepackage{amssymb}
\usepackage{wrapfig}
\usepackage{tabularx}
\usepackage{textcomp}
\usepackage{pifont}
\usepackage{xspace}
\usepackage{bbm}

\newcommand{\ours}{$\text{HiR}^{2}$\xspace}
\definecolor{lightgray}{rgb}{0.7,0.7,0.7}

\definecolor{mygreen}{HTML}{009900}
\definecolor{myred}{HTML}{CC0000}
\definecolor{mygray}{HTML}{666666}

\icmltitlerunning{Learning Taxonomic Trees with Hierarchical Representation Regularization for Large Multimodal Models}

\begin{document}

\twocolumn[
  \icmltitle{Learning Taxonomic Trees with Hierarchical Representation \\ Regularization for Large Multimodal Models}



  \icmlsetsymbol{equal}{*}

  \begin{icmlauthorlist}
    \icmlauthor{Hulingxiao He}{yyy}
    \icmlauthor{Zhi Tan}{yyy}
    \icmlauthor{Yuxin Peng}{yyy}
  \end{icmlauthorlist}

  \icmlaffiliation{yyy}{Wangxuan Institute of Computer Technology, Peking University}

  \icmlcorrespondingauthor{Yuxin Peng}{pengyuxin@pku.edu.cn}

  \icmlkeywords{Machine Learning, ICML}

  \vskip 0.3in
]



\printAffiliationsAndNotice{}  

\begin{abstract}

Taxonomies provide key information about the semantic relationships between concepts and the inherent organization of vision and language. Despite their impressive capabilities, large multimodal models (LMMs) often lack taxonomic knowledge, leading to low hierarchical visual recognition (HVR) consistency. These models typically only rely on language modeling objectives during fine-tuning and lack explicit taxonomy-aware regularization. To address this, we propose \textit{Hierarchical Representation Regularization (\ours)}, a simple plug-and-play regularizer that improves hierarchical consistency in LMMs. Specifically, we introduce a semantic-aware visual tree construction framework that extracts coarse-to-fine visual features from intermediate LLM layers guided by textual cues. The regularizer combines two complementary objectives: a taxonomic entailment loss that enforces hierarchy via hyperbolic entailment cones in the Lorentz model, and a discriminative dispersive loss that promotes angular separation of semantically similar embeddings on the unit sphere \textit{without disturbing the radial hierarchical structure}. Extensive experiments demonstrate that \ours~effectively captures taxonomic structures across diverse LMMs and fine-tuning methods. Code is available at \href{https://github.com/PKU-ICST-MIPL/HiR2_ICML2026}{https://github.com/PKU-ICST-MIPL/HiR2\_ICML2026}.

\end{abstract}

\section{Introduction}
\label{sec:intro}

\begin{figure}[t]
    \centering
    \includegraphics[width=0.98\linewidth]{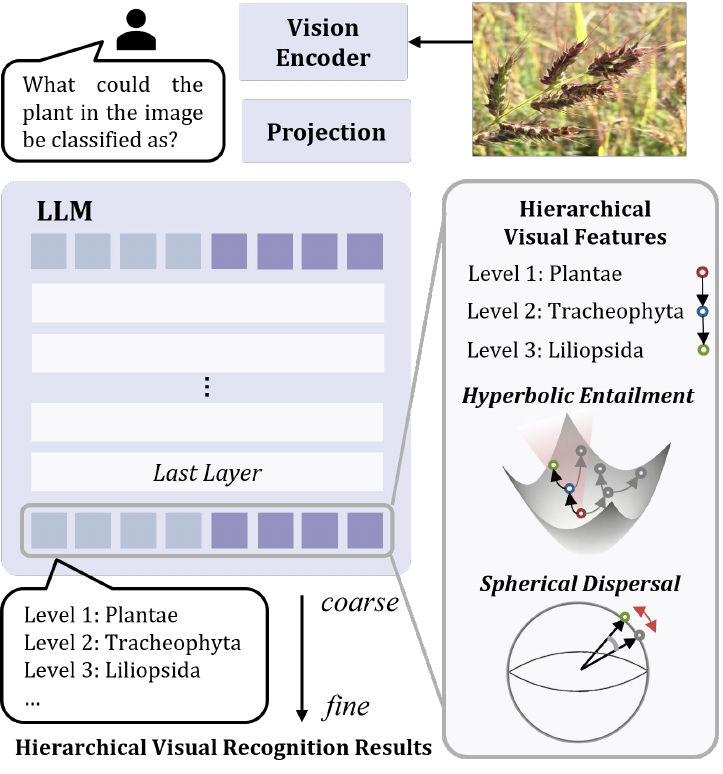}
    \caption{Hierarchical Representation Regularization encourages the intermediate representations to learn tree-like structures according to taxonomy beyond language modeling loss, thereby enhancing hierarchical visual recognition capability.}
    \label{fig:motivation}
\end{figure}

The real world is not binary, but governed by taxonomies. 
Taxonomies encode rich semantic relations between concepts and reflect the inherent organization of vision and language. For example, hierarchical visual recognition (HVR) aims at categorizing real-world concepts that are naturally organized at multiple levels of abstraction. 
These relations form tree-structured hierarchies, where general concepts such as ``plants'' reside near the root, while more specific categories such as ``trees'' or ``flowers'' appear at deeper levels. 
Effectively modeling such structures is fundamental to hierarchical understanding of the visual world.

While existing large multimodal models (LMMs)~\cite{Qwen2.5-VL,finedefics,deepperception} have demonstrated strong performance in fine-grained visual recognition (FGVR)~\cite{peng2025survey}, 
they remain weak in HVR and often fail to maintain hierarchical consistency~\cite{tan2025vision}. 
This limitation largely stems from their training objectives: LMMs are primarily optimized with a language modeling loss for next-token prediction, without any explicit \textit{taxonomy-aware regularization} imposed on the visual representations used for language generation. This paradigm contrasts sharply with prior work on discriminative vision~\cite{hcast2025park} or vision-language models~\cite{stevens2024bioclip,pal2024compositional,wei2025modality}, where hierarchical representation learning has been a central research topic for enhancing the prediction consistency.

Motivated by hierarchical representation learning, we are \textit{the first} to explore its integration into LMMs and propose \textit{Hierarchical Representation Regularization} (\ours), a flexible and plug-and-play framework that injects taxonomic knowledge into LMMs without introducing additional learnable parameters~\cite{xiao2026survey}. As illustrated in Figure~\ref{fig:motivation}, the key idea is to complement the standard language modeling objective with auxiliary losses that explicitly regularize internal visual representations to follow taxonomic tree structures. 
This formulation introduces two main challenges: 
(1) extracting hierarchical visual features with coarse-to-fine semantics from LMMs, and 
(2) designing learning objectives that faithfully encode taxonomic relations.

To address the first challenge, we develop a semantic-aware framework that leverages hidden states from the last layer of the LLM to construct hierarchical visual features. 
Specifically, a non-parametric cross-attention module is introduced, where textual features at each hierarchy level act as queries and visual token embeddings serve as keys and values, producing hierarchy-specific visual representations. 
These representations naturally form a semantic-aware visual tree with coarse-to-fine semantics.

To address the second challenge, we introduce two complementary learning objectives to structure the taxonomic tree. 
The first is a \emph{taxonomic entailment loss}, which enforces geometric containment constraints in the hyperbolic Lorentz model, ensuring that child concepts (e.g., \emph{Flower}) lie within the entailment cone of their parents (e.g., \emph{Plant}). 
Hyperbolic space is particularly suitable for this purpose due to its exponential volume growth. 
The second is a \emph{discriminative dispersive loss}, which increases angular separation between visually ambiguous sibling categories while preserving the radial hierarchy, thereby improving fine-grained discriminability without disrupting the taxonomic structure. 
The training procedure follows standard LMM fine-tuning practices, only adding additional regularization terms that incur little computational overhead. With its minimalist and self-contained design, \ours~demonstrates that injecting taxonomy knowledge via hierarchical representation learning can substantially benefit LMMs.

We evaluate the effectiveness and generality of \ours~through extensive experiments. 
Results show consistent improvements over strong LMM baselines such as Qwen2.5-VL~\cite{Qwen2.5-VL}, Qwen2-VL~\cite{Qwen2-vl}, Intern3.5-VL-1B~\cite{wang2025internvl3_5}, and LLaVA-OV-1.5-4B\cite{LLaVA-OneVision-1.5}, 
across both supervised fine-tuning (SFT) and dynamic fine-tuning (DFT)~\cite{wu2025generalization} methods on various taxonomies like iNaturalist-2021 (iNat21)~\cite{inat2021} and CUB-200-2011 (CUB-200)~\cite{cub200}. These findings highlight the simplicity, effectiveness, and broad applicability of \ours~for learning taxonomic structures in LMMs. 
Our contributions are summarized as follows:

(1) This work presents the first principled study that introduces hierarchical representation learning as a regularization mechanism for large multimodal models beyond standard language modeling objectives.

(2) A semantic-aware visual tree construction framework is proposed to extract hierarchical visual representations with coarse-to-fine semantics from intermediate layers of large multimodal models.

(3) Two complementary objectives are developed: a taxonomic entailment loss that encodes hierarchical priors in hyperbolic space, and a dispersive loss that improves fine-grained discrimination via angular separation while preserving the radial hierarchical structure.

\section{Related Work}

\noindent\textbf{Hierarchical Visual Recognition.}
HVR~\cite{silla2011survey,kosmopoulos2015evaluation} plays a central role in understanding both visual~\cite{por,park2024learning,zeng2024learning,sinha2024learning,chen2022label,parkvisually} and language concepts~\cite{zhou2020hierarchy,wang2022incorporating,zhou2025novel,he2024language}. Recent studies reveal that CLIP-style models~\cite{clip} often lack taxonomic consistency~\cite{protect,geng2023hiclip}. To address this, \cite{protect} evaluate CLIP across multiple granularity levels and propose hierarchy-consistent prompt tuning, while others enhance CLIP via hyperbolic embeddings~\cite{pal2024compositional} or graph-based learning~\cite{xia2023hgclip}. \cite{novack2023chils} further show that hierarchical supervision can improve zero-shot classification.
Beyond CLIP, \cite{vlm_img_bad} first identify the limitations of LMMs in FGVR, later extended by \cite{liu2024revisiting}. \cite{he2025analyzing} attribute this issue to insufficient exposure to class names during pretraining. Subsequent evaluations consider both closed-set~\cite{yu2025benchmarking,geigle2024african,he2026fine} and open-world settings~\cite{conti2025large,he2026fine}, with \cite{snaebjarnarson2025taxonomy} advocating taxonomic similarity over exact label matching. Most recently, \cite{tan2025vision} explicitly analyze the hierarchical understanding of LMMs, highlighting persistent challenges in hierarchical consistency and leaf-level accuracy. \cite{he2026taxonomy} propose representation alignment to inject taxonomic knowledge from discriminative models into generative LMMs.

\noindent\textbf{Learning on Hyperbolic Manifolds.}
Hyperbolic manifolds provide an effective geometry for modeling hierarchical structures, motivating a growing body of work on hyperbolic neural networks that incorporate non-Euclidean operations for representation learning~\citep{guo2022clipped,shimizu2020hyperbolic,he2025lorentzian,malik2025hyperdefender,skopek2019mixed,gao2021curvature,yu2025hyperbolic,fan2025curvature}. They have been successfully applied across graphs~\citep{fu2023hyperbolic,fu2024hyperbolic,malik2025hyperdefender}, text~\citep{he2025helm}, images~\citep{Wang_2024_CVPR,franco2023hyperbolic_al,li2025hyperbolic,gao2023exploring,li2025geometry} and videos~\citep{long2020searching,hong2023curved}.
Recent advances further extend hyperbolic learning to multimodal settings by combining entailment-aware objectives with CLIP-style training~\citep{ramasinghe2024accept,desai2023hyperbolic,palcompositional,wang2024cliploss}, and by scaling hyperbolic representations to large vision–language models~\citep{mandica2024hyperbolic,wei2025modality}.

\noindent\textbf{Representation Learning as Auxiliary Tasks.}
Beyond standard pre-training and fine-tuning, representation learning is often incorporated as an auxiliary objective jointly optimized with the main task~\cite{ye2022unsupervised}. Supervised contrastive learning~\cite{khosla2020supervised} augments classification with contrastive objectives, while SLIP~\cite{mu2022slip} extends CLIP~\cite{radford2021learning} by adding a parallel self-supervised signal. In image generation, REPA~\cite{repa} aligns intermediate generative features with those of a frozen encoder, later improved by SARA~\cite{chen2025sara} through structural and adversarial alignment, and extended to multimodal settings by SoftREPA~\cite{softrepa}. For LMMs, VIRAL~\cite{viral} aligns internal visual representations with pretrained vision encoders to inject complementary visual knowledge, while JARVIS~\cite{jarvis} employs a masked predictive objective to align predicted intermediate representations with those of a target encoder using a single context block.

\begin{figure*}
    \centering
    \includegraphics[width=0.98\linewidth]{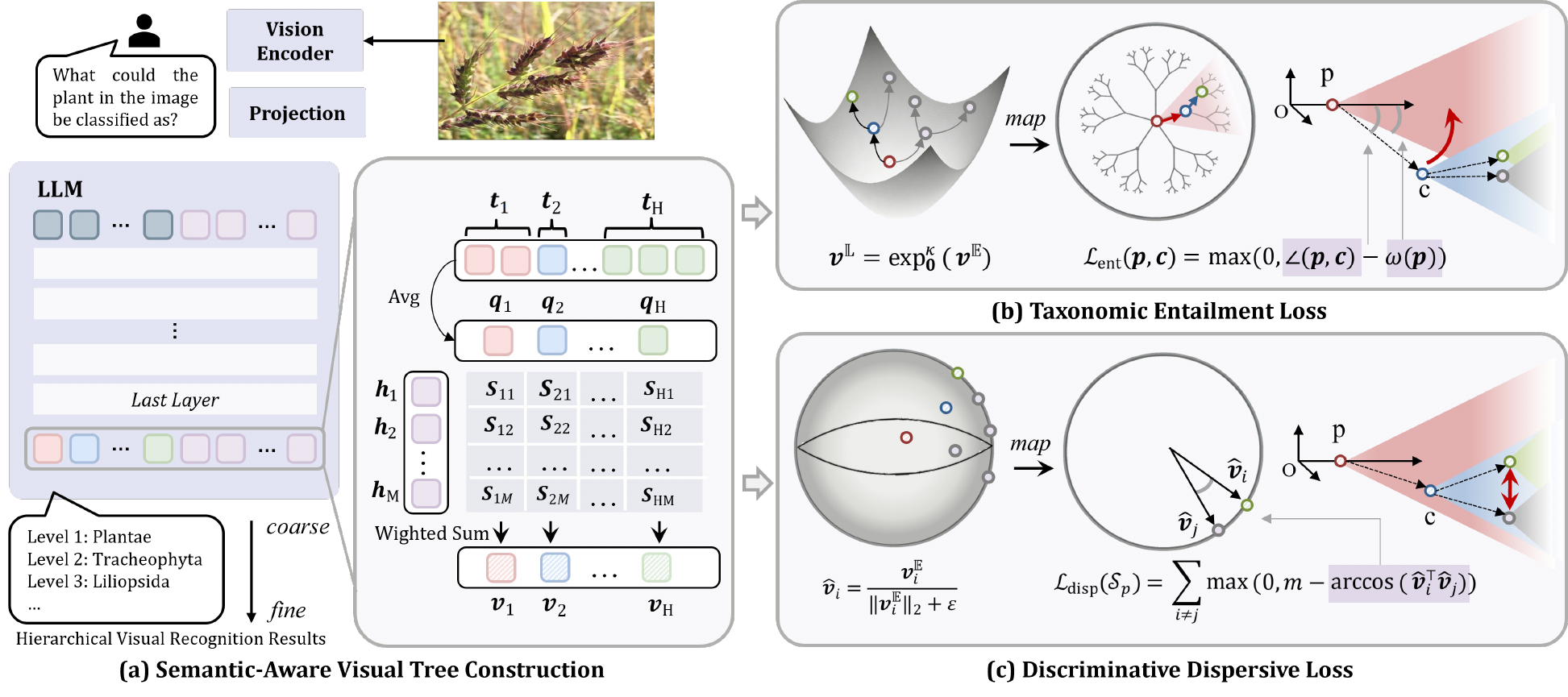}
    \caption{Overview of the \ours framework. (a) Constructing hierarchical visual features from the last LLM layer. (b) Optimizing with a taxonomic entailment loss to enforce the hierarchy. (c) Optimizing with a discriminative dispersive loss for distinguishing semantically-similar categories.}
    \label{fig:pipeline}
\end{figure*}

\section{Problem Setting and Preliminaries}

In this section, we introduce the problem setting and the preliminaries considered in this paper.

\textbf{Problem Setting.}
Conventional visual recognition typically assumes a flat label space, where each image $x \in \mathcal{X}$ is assigned a single label $y \in \mathcal{Y}$. However, real-world visual concepts are often structured hierarchically, with labels organized in a taxonomy $\mathcal{T}=(\mathcal{Y},\mathcal{E})$~\cite{parkvisually,protect,por,xia2023hgclip}, such as a tree or a directed acyclic graph (DAG). Each directed edge $(y_i, y_j)\in\mathcal{E}$ encodes a parent--child relationship, where $y_i$ is the parent of $y_j$. In hierarchical visual recognition (HVR), the goal is to predict not only a leaf label $y \in \mathcal{Y}_{\text{leaf}}$, but also its full ancestral path $(y_0, y_1, \ldots, y_L)$ from root to leaf.

\noindent \textbf{Hyperbolic Manifold.}
Unlike Euclidean spaces with zero curvature, a hyperbolic manifold is a smooth Riemannian manifold with constant negative curvature $-\kappa$ ($\kappa>0$)~\citep{lee2006riemannian}. 
Following prior work~\citep{cannon1997hyperbolic}, we adopt the \emph{Lorentz model} for hyperbolic geometry due to its computational efficiency and numerical stability.
Formally, the $d$-dimensional Lorentz model is defined as
\begin{equation}
\mathbb{L}^{d,\kappa}
=
\left\{
\boldsymbol{p} \in \mathbb{R}^{d+1}
\;\middle|\;
\langle \boldsymbol{p}, \boldsymbol{p} \rangle_{\mathbb{L}} = -\frac{1}{\kappa},
\; p_0 > 0
\right\},
\end{equation}
which corresponds to the upper sheet of a two-sheeted hyperboloid embedded in $(d\!+\!1)$-dimensional Minkowski space.

Each point $\boldsymbol{p} \in \mathbb{L}^{d,\kappa}$ is represented as
$\boldsymbol{p} = [p_0, \tilde{\boldsymbol{p}}]$,
where $p_0 \in \mathbb{R}$ is the \emph{time-like} component and
$\tilde{\boldsymbol{p}} \in \mathbb{R}^d$ is the \emph{space-like} component.
The Lorentzian inner product
$\langle \cdot, \cdot \rangle_{\mathbb{L}}$
between two points $\boldsymbol{p}=[p_0,\tilde{\boldsymbol{p}}]$ and
$\boldsymbol{q}=[q_0,\tilde{\boldsymbol{q}}]$ is defined as
\begin{equation}
\langle \boldsymbol{p}, \boldsymbol{q} \rangle_{\mathbb{L}}
=
- p_0 q_0
+
\langle \tilde{\boldsymbol{p}}, \tilde{\boldsymbol{q}} \rangle_{\mathbb{E}},
\end{equation}
where $\langle \cdot, \cdot \rangle_{\mathbb{E}}$ denotes the Euclidean inner product.
The induced Lorentzian norm is
$\|\boldsymbol{p}\|_{\mathbb{L}} = \sqrt{|\langle \boldsymbol{p}, \boldsymbol{p} \rangle_{\mathbb{L}}|}$.

\noindent \textbf{Distance.}
The geodesic (shortest-path) distance between two points
$\boldsymbol{p}, \boldsymbol{q} \in \mathbb{L}^{d,\kappa}$
is given by
\begin{equation}
d_{\mathbb{L}}(\boldsymbol{p}, \boldsymbol{q})
=
\frac{1}{\sqrt{\kappa}}
\operatorname{arccosh}
\!\left(
-\kappa \langle \boldsymbol{p}, \boldsymbol{q} \rangle_{\mathbb{L}}
\right).
\end{equation}

\noindent \textbf{Tangent Space.}
Each point $\boldsymbol{p} \in \mathbb{L}^{d,\kappa}$ is associated with a
$d$-dimensional tangent space $T_{\boldsymbol{p}}\mathbb{L}^{d,\kappa}$,
which is a Euclidean vector space providing a local linear approximation.
Any ambient vector $\boldsymbol{u} \in \mathbb{R}^{d+1}$ can be projected onto
$T_{\boldsymbol{p}}\mathbb{L}^{d,\kappa}$ via
\begin{equation}
\operatorname{proj}^{\kappa}_{\boldsymbol{p}}(\boldsymbol{u})
=
\boldsymbol{u}
+
\kappa \boldsymbol{p}
\langle \boldsymbol{p}, \boldsymbol{u} \rangle_{\mathbb{L}} .
\end{equation}

\noindent \textbf{Exponential Map.}
The exponential map
$\exp^{\kappa}_{\boldsymbol{p}} : T_{\boldsymbol{p}}\mathbb{L}^{d,\kappa} \rightarrow \mathbb{L}^{d,\kappa}$
projects a tangent vector $\boldsymbol{v} \in T_{\boldsymbol{p}}\mathbb{L}^{d,\kappa}$
onto the manifold:
\begin{equation}
\begin{aligned}
\exp^{\kappa}_{\boldsymbol{p}}(\boldsymbol{v})
=
\cosh\!\left(\sqrt{\kappa}\,\|\boldsymbol{v}\|_{\mathbb{L}}\right)\boldsymbol{p}
+
\frac{\sinh\!\left(\sqrt{\kappa}\,\|\boldsymbol{v}\|_{\mathbb{L}}\right)}
{\sqrt{\kappa}\,\|\boldsymbol{v}\|_{\mathbb{L}}}
\,\boldsymbol{v}.
\end{aligned}
\end{equation}
In particular, the exponential map at the origin
$\mathbf{0} = (\sqrt{1/\kappa}, 0, \ldots, 0)^{\top}$
is used to embed Euclidean features into hyperbolic space.

\noindent \textbf{Logarithmic Map.}
Conversely, the logarithmic map
$\log^{\kappa}_{\boldsymbol{q}} : \mathbb{L}^{d,\kappa} \rightarrow T_{\boldsymbol{q}}\mathbb{L}^{d,\kappa}$
maps a point $\boldsymbol{p} \in \mathbb{L}^{d,\kappa}$ to the tangent space at
$\boldsymbol{q}$:
\begin{equation}
\begin{aligned}
\log^{\kappa}_{\boldsymbol{q}}(\boldsymbol{p})
=
\frac{
\operatorname{arccosh}\!\left(-\kappa \langle \boldsymbol{q}, \boldsymbol{p} \rangle_{\mathbb{L}}\right)
}{
\sqrt{(\kappa \langle \boldsymbol{q}, \boldsymbol{p} \rangle_{\mathbb{L}})^2 - 1}
}
\operatorname{proj}^{\kappa}_{\boldsymbol{q}}(\boldsymbol{p}).
\end{aligned}
\end{equation}

\textbf{Hyperbolic Entailment Cones.}
Hyperbolic entailment cones associate each point
$\boldsymbol{p} = [p_0, \tilde{\boldsymbol{p}}] \in \mathbb{L}^{d,\kappa}$ 
with a cone-shaped region $\omega(\boldsymbol{p})$, such that any
$\boldsymbol{c} = [c_0, \tilde{\boldsymbol{c}}] \in \omega(\boldsymbol{p})$
is interpreted as a child of $\boldsymbol{p}$.
The cone's half-aperture is
\begin{equation}
\omega(\boldsymbol{p}) = \arcsin\!\left(
\frac{2\gamma}{\sqrt{\kappa}\,\|\tilde{\boldsymbol{p}}\|}
\right),
\end{equation}
where $\gamma = 0.1$ in our experiments.

\section{Methodology}

Motivated by the potential of representation learning for improving hierarchical consistency in LMMs, we propose \textit{Hierarchical Representation Regularization} (\ours), a plug-and-play framework that explicitly injects taxonomic structure into the intermediate visual representations of LMMs. As illustrated in Figure~\ref{fig:pipeline}, \ours constructs hierarchy-aware visual features from internal hidden states, and enforces both hierarchical entailment and sibling-level discriminability through geometry-aware regularization objectives. \ours~can be easily computed without introducing additional learning parameters, as shown in Algorithm~\ref{algorithm}.

\subsection{Semantic-aware Visual Tree Construction}

 Motivated by the cross-modal information flow mechanism of LMMs \cite{zhang2025cross}, we leverage hidden states from intermediate (or last) layers to construct hierarchical visual features corresponding to different levels of the taxonomic tree.

 \textbf{Hierarchical Visual Feature Extraction.} We denote the textual embedding sequence corresponding to different levels of categories as
$[\boldsymbol{T}_1; \boldsymbol{T}_2; \ldots; \boldsymbol{T}_H]$,
where $H$ denotes the depth of the textual hierarchy. For the $i$-th level,
\begin{equation}
\small
\boldsymbol{T}_i = [\boldsymbol{t}_i(1); \boldsymbol{t}_i(2); \ldots; \boldsymbol{t}_i(N_i)],
\end{equation}
and $N_i$ is the token length of the $i$-th level category name.

Let $\boldsymbol{H} = [\boldsymbol{h}_1; \boldsymbol{h}_2; \ldots; \boldsymbol{h}_M] \in \mathbb{R}^{M \times D}$ denote the hidden states corresponding to image tokens at the same transformer layer, where $M$ is the number of image tokens and $D$ is the embedding dimension. We first obtain a level-wise textual query embedding by average pooling over tokens belonging to the same hierarchy level:
\begin{equation}
\small
\boldsymbol{q}_i = \frac{1}{N_i} \sum_{j=1}^{N_i} \boldsymbol{t}_i(j), \quad i = 1, \ldots, H.
\end{equation}

To extract visual features aligned with different hierarchy levels, we design a nonparametric cross-attention mechanism where each textual query $\boldsymbol{q}_i$ attends to image token features. Specifically, the attention weights over image tokens are computed as
\begin{equation}
\small
\mathcal{S}_{ij} = 
\frac{
\exp\left( \boldsymbol{q}_i^\top \boldsymbol{h}_j / \sqrt{D} \right)
}{
\sum\limits_{k \in \mathcal{I}}
\exp\left( \boldsymbol{q}_i^\top \boldsymbol{h}_k / \sqrt{D} \right)
},
\end{equation}
where $\mathcal{I}$ denotes the set of positions corresponding to image tokens, ensuring that attention is restricted to the visual modality.

The hierarchical visual embedding for the $i$-th level is then obtained as a weighted aggregation of image token features:
\begin{equation}
\small
\boldsymbol{v}_i = \sum_{j \in \mathcal{I}} \mathcal{S}_{ij} \boldsymbol{h}_j,
\quad i = 1, \ldots, H.
\end{equation}

As a result, we obtain a hierarchy-aware visual feature set
$\{\boldsymbol{v}_i^\mathbb{E}\}_{i=1}^H$, where each $\boldsymbol{v}_i^\mathbb{E} \in \mathbb{R}^D$ captures hierarchy-aligned visual semantics in Euclidean space, serving as a geometry-agnostic intermediate representation before hierarchical regularization.

\textbf{Curvature-aware Scaling.}
Before mapping Euclidean visual features into hyperbolic space, we apply a \emph{curvature-aware scaling} operation to normalize embedding magnitudes at the batch level.
Given Euclidean visual embeddings $\boldsymbol{v}_i^\mathbb{E} \in \mathbb{R}^D$, we first compute their $\ell_2$ norms and estimate the batch-wise average norm
\begin{equation}
\small
\bar{r}
=
\frac{1}{|\mathcal{B}|}
\sum_{i \in \mathcal{B}}
\|\boldsymbol{v}_i^\mathbb{E}\|_2 ,
\end{equation}
where $\mathcal{B}$ denotes the set of embeddings within the current batch. We then rescale all embeddings using a shared scaling factor
\begin{equation}
\small
s
=
\frac{\rho}{\sqrt{\kappa}\,(\bar{r} + \varepsilon)},
\end{equation}
where $\kappa$ is the target hyperbolic curvature, $\rho \in (0,1)$ is a predefined target ratio, and $\varepsilon$ is a small constant for numerical stability.
The scaled embeddings are given by
\begin{equation}
\small
\tilde{\boldsymbol{v}}_i^\mathbb{E}
=
s \cdot \boldsymbol{v}_i^\mathbb{E}.
\end{equation}

Importantly, the average norm $\bar{r}$ is detached from the computational graph, so the scaling factor does not introduce additional gradients.
This batch-level normalization aligns the expected embedding radius with the curvature-dependent scale $1/\sqrt{\kappa}$, stabilizing subsequent exponential mapping into the Lorentz model and preventing excessive norm growth during training.

\textbf{Exponential Map.} After Curvature-aware Scaling, we map all
Euclidean features into the hyperbolic embeddings of Lorentz model. This transition is achieved via the exponential map
\cite{nickel2018learning} at the origin $\mathbf{0}$, denoted as
$\exp^{\kappa}_{\mathbf{0}} : T_{\mathbf{0}}\mathbb{L}^{d,\kappa} \rightarrow \mathbb{L}^{d,\kappa}$.
For ease of expression, $\boldsymbol{v}^\mathbb{E}$ is used to denote any Euclidean
visual features, \emph{e.g.}, $[\boldsymbol{v}_1^\mathbb{E}; \boldsymbol{v}_2^\mathbb{E}; \ldots; \boldsymbol{v}_H^\mathbb{E}]$.
Its corresponding hyperbolic embedding
$\boldsymbol{v}^\mathbb{L} \in \mathbb{L}^{d,\kappa}$ is computed as $\boldsymbol{v}^\mathbb{L}
= \exp^{\kappa}_{\mathbf{0}}(\boldsymbol{v}^\mathbb{E})$.

\subsection{Taxonomic Entailment Loss}

To explicitly encode the hierarchical structure of taxonomic trees, we devise a \emph{taxonomic entailment loss} $\mathcal{L}_\mathrm{ent}$ defined in hyperbolic space. Unlike Euclidean or spherical geometries whose volume growth is polynomial or constant with respect to radius, hyperbolic space exhibits exponential volume expansion, closely matching the branching property of tree-structured data. This geometric property allows hierarchical depth to be naturally encoded by radial distance, making hyperbolic space particularly suitable for modeling parent-child relations in deep taxonomies.

Formally, this loss leverages
hyperbolic entailment cones \cite{ganea2018hyperbolic,desai2023hyperbolic},
wherein a parent concept $p$ geometrically entails its child concept $q$ by
constraining them to lie within its cone-shaped region. We apply this loss to
preserve both semantic hierarchy and conceptual hierarchy. 
For each parent--child pair $(\boldsymbol{p}, \boldsymbol{c})$,
we penalize child embeddings outside the cone:
\begin{equation}
\small
\mathcal{L}_{\mathrm{ent}}(\boldsymbol{p}, \boldsymbol{c})
= \max\bigl(0,\; \angle(\boldsymbol{p},\boldsymbol{c}) - \omega(\boldsymbol{p}) \bigr),
\end{equation}
where
\begin{equation}
\small
\angle(\boldsymbol{p},\boldsymbol{c})
= \arccos\!\left(
\frac{p_0 + \kappa \langle \boldsymbol{p}, \boldsymbol{c} \rangle_{\mathbb{L}} c_0}
{\|\tilde{\boldsymbol{c}}\| \sqrt{(\kappa \langle \boldsymbol{p}, \boldsymbol{c} \rangle_{\mathbb{L}})^2 - 1}}
\right).
\end{equation}
The total entailment loss across hierarchy levels is
\begin{equation}
\small
\mathcal{L}_\mathrm{ent} = \sum_{i=1}^{H-1}
\mathcal{L}_{\mathrm{ent}}\!\left(
\boldsymbol{v}_i^\mathbb{L}, \boldsymbol{v}_{i+1}^\mathbb{L}
\right).
\end{equation}

\begin{algorithm}[t]
\caption{LMM Fine-tuning + \ours}
\label{algorithm}
\begin{algorithmic}[1]
\REQUIRE Outputs $\texttt{outputs}$, labels $\texttt{labels}$, image hidden states $\boldsymbol{H}$, text hierarchy $\{\boldsymbol{T}_i\}$, taxonomy $\mathcal{T}$
\ENSURE Total loss $\mathcal{L}_{\mathrm{total}}$

\STATE Compute language modeling loss $\mathcal{L}_{\mathrm{LM}}$

\FOR{each selected image--text layer}
    \STATE Construct hierarchical Euclidean visual features $\{\boldsymbol{v}_i^{\mathbb{E}}\}_{i=1}^{H}$ via cross-attention
    \STATE Apply curvature-aware scaling to $\boldsymbol{v}_i^{\mathbb{E}}$
    \STATE Map to hyperbolic space $\boldsymbol{v}_i^{\mathbb{L}}=\exp^{\kappa}_{\mathbf{0}}(\boldsymbol{v}_i^{\mathbb{E}})$
    \STATE Accumulate taxonomic entailment loss $\mathcal{L}_{\mathrm{ent}}$ over $(\boldsymbol{v}_i^{\mathbb{L}},\boldsymbol{v}_{i+1}^{\mathbb{L}})$
    \STATE Project $\boldsymbol{v}_i^{\mathbb{E}}$ to angular components $\hat{\boldsymbol{v}}_i$
    \STATE Accumulate parent-aware dispersive loss $\mathcal{L}_{\mathrm{disp}}$ with memory banks
\ENDFOR

\STATE Apply scheduled activation $\alpha$ to $\mathcal{L}_{\mathrm{disp}}$

\STATE $\mathcal{L}_{\mathrm{total}}
= \mathcal{L}_{\mathrm{LM}}
+ \lambda_{\mathrm{ent}}\mathcal{L}_{\mathrm{ent}}
+ \lambda_{\mathrm{disp}}\mathcal{L}_{\mathrm{disp}}$
\end{algorithmic}
\end{algorithm}

\subsection{Discriminative Dispersive Loss}

While the hyperbolic taxonomic entailment loss effectively preserves hierarchical depth, it does not explicitly enforce discrimination among sibling categories under the same parent. In deep hierarchies, such sibling embeddings may collapse along similar directions, leading to ambiguous fine-grained representations. 

To address this issue without interfering with hyperbolic radial hierarchy, we introduce a \emph{discriminative dispersive loss} defined on a spherical space, where embeddings are constrained to lie on a unit hypersphere and discrimination is governed purely by angular separation. The fixed radius and bounded volume of spherical space make it particularly suitable for modeling sibling-level diversity while remaining complementary to hyperbolic hierarchy modeling.

\textbf{Spherical Angular Separation.}
By projecting Euclidean visual embeddings onto a unit hypersphere, we isolate their angular components by $\hat{\boldsymbol{v}}_i
= \frac{\boldsymbol{v}_i^\mathbb{E}}{\|\boldsymbol{v}_i^\mathbb{E}\|_2 + \varepsilon}$.

This spherical normalization removes radial degrees of freedom, ensuring that the dispersive objective depends solely on angular relationships and does not introduce implicit hierarchical ordering.

\textbf{Sibling-only Repulsion.}
On the resulting unit hypersphere, we define sibling-level repulsion purely in terms of angular distance. Formally, for parent $p$, let
$\mathcal{S}_p = \{ \hat{\boldsymbol{v}}_i \mid \text{parent}(\boldsymbol{v}_i) = p \}$.
We encourage angular separation among siblings with a margin-based objective:
\begin{equation}
\small
\mathcal{L}_{\mathrm{disp}}(\mathcal{S}_p)
= \sum_{i \neq j} \max\!\bigl(0, m - \arccos(\hat{\boldsymbol{v}}_i^\top \hat{\boldsymbol{v}}_j)\bigr),
\end{equation}
while ignoring embeddings from different parents.

\textbf{Parent-aware Memory Bank.}
To enhance stability and long-term diversity, each parent $p$ maintains a memory bank $\mathcal{M}_p$ of past embeddings. After angular projection, interactions with memory extend the loss:
\begin{equation}
\small
\small
\mathcal{L}_{\mathrm{disp}}(\mathcal{S}_p, \mathcal{M}_p)
= \!\!\!\sum_{\hat{\boldsymbol{v}} \in \mathcal{S}_p} \sum_{\hat{\boldsymbol{u}} \in \mathcal{M}_p} 
\max\!\bigl(0, m - \arccos(\hat{\boldsymbol{v}}^\top \hat{\boldsymbol{u}})\bigr).
\end{equation}

\textbf{Scheduled Activation.}
The dispersive loss is gradually activated with a linearly increasing factor $\alpha \in [0,1]$:
\begin{equation}
\small
\mathcal{L}_{\mathrm{disp}} = \alpha \sum_p \mathcal{L}_{\mathrm{disp}}(\mathcal{S}_p, \mathcal{M}_p),
\end{equation}
aggregating over all parents in the taxonomy.

\textbf{Total Loss.}
Finally, the overall objective combines language modeling and the two hierarchical regularizers:
\begin{equation}
\small
\mathcal{L}_{\mathrm{total}}
= \mathcal{L}_{\mathrm{LM}} + \lambda_\mathrm{ent}\mathcal{L}_{\mathrm{ent}} + \lambda_\mathrm{disp}\mathcal{L}_{\mathrm{disp}}.
\end{equation}

\begin{table}[t]
\centering
\caption{
Variants of the dispersive loss. 
}
\small
\begin{tabular}{c l l}
\toprule
Variant &
$\delta(\cdot,\cdot)$ &
Space / Reference \\
\midrule
I &
$d_{\mathbb{H}}(\boldsymbol{z}_i,\boldsymbol{z}_j)$ &
$\mathbb{H}$ \\

II &
$d_{\mathrm{ang}}\!\left(
\log_{\mathbf{0}}(\boldsymbol{z}_i),
\log_{\mathbf{0}}(\boldsymbol{z}_j)
\right)$ &
$T_{\mathbf{0}}\mathbb{H}$ \\

III &
$d_{\mathrm{ang}}\!\left(
\log_{\boldsymbol{p}}(\boldsymbol{z}_i),
\log_{\boldsymbol{p}}(\boldsymbol{z}_j)
\right)$ &
$T_{\boldsymbol{p}}\mathbb{H}$ \\

IV &
$d_{\mathrm{ang}}\!\left(
\boldsymbol{z}_i-\boldsymbol{p},
\boldsymbol{z}_j-\boldsymbol{p}
\right)$ &
$\mathbb{R}^D$ \\

Ours &
$d_{\mathrm{ang}}\!\left(
\hat{\boldsymbol{v}}_i,
\hat{\boldsymbol{v}}_j
\right),\;
\|\hat{\boldsymbol{v}}\|_2=1$ &
$\mathbb{S}^{D-1}$ \\
\bottomrule
\end{tabular}
\label{tab:ablation_disp_variants}
\end{table}

\begin{table*}[ht]
\centering
\caption{Performance evaluation on iNat-Plant and iNat-Animal with different hierarchical distributions. The best results are in \sethlcolor{lightpink}\hl{green}.}
\vspace{-0.35em}
\resizebox{1.0\linewidth}{!}{
\begin{tabular}{ll|lcccccccc}
\toprule
\multirow{2}{*}[-0.5ex]{{Dataset}} & \multirow{2}{*}[-0.5ex]{\shortstack{Hierarchy\\Distribution}}
 & \multirow{2}{*}[-0.5ex]{{Methods}}  
 & \multicolumn{4}{c}{{Base}} 
 & \multicolumn{4}{c}{{Novel}}  
 \\ 
\cmidrule(l){4-7} \cmidrule(l){8-11} 
&  
&
& {HCA} $\uparrow$
& {POR} $\uparrow$ 
& {S-POR} $\uparrow$ 
& {TOR} $\uparrow$
& {HCA} $\uparrow$
& {POR} $\uparrow$ 
& {S-POR} $\uparrow$ 
& {TOR} $\uparrow$
\\ 
\midrule

\multirow{4}{*}[-0ex]{\shortstack{iNat-\\Plant}} & \multirow{4}{*}[-0ex]{\shortstack{5-14-85-286\\-1702-4271}}

 & SFT  
    & 9.98  & 63.63 & 46.01 & 45.69 & 10.86  & 63.74 & 46.15 & 45.16
   \\
   
& & SFT+Ours 
    & \sethlcolor{lightpink}\hl{11.90} \up{1.92} 
    & \sethlcolor{lightpink}\hl{66.65} \up{3.01}
    & \sethlcolor{lightpink}\hl{50.87} \up{4.86}
    & \sethlcolor{lightpink}\hl{49.76} \up{4.07}
    & \sethlcolor{lightpink}\hl{13.06} \up{2.20}
    & \sethlcolor{lightpink}\hl{67.16} \up{3.42}
    & \sethlcolor{lightpink}\hl{50.69} \up{4.54}
    & \sethlcolor{lightpink}\hl{49.69} \up{4.53}
   \\ 

& & DFT 
    & 8.15 & 62.45 & 40.47 & 41.98 & 9.32 & 63.26 & 41.35 & 42.64
   \\

& & {DFT+Ours}
    & \sethlcolor{lightpink}\hl{11.66} \up{3.51}
    & \sethlcolor{lightpink}\hl{66.07} \up{3.62}
    & \sethlcolor{lightpink}\hl{49.83} \up{9.36}
    & \sethlcolor{lightpink}\hl{48.81} \up{6.83}
    & \sethlcolor{lightpink}\hl{12.87} \up{3.55}
    & \sethlcolor{lightpink}\hl{66.72} \up{3.46}
    & \sethlcolor{lightpink}\hl{50.12} \up{8.77}
    & \sethlcolor{lightpink}\hl{49.10} \up{6.46}
   \\ \cmidrule(r){1-11}

\multirow{4}{*}[-0ex]{\shortstack{iNat-\\Animal}} &
\multirow{4}{*}[-0ex]{\shortstack{6-27-152-715\\-2988-5388}}

& SFT  
    & 14.06  & 71.41 & 60.40 & 58.03 & 15.74 & 72.12 & 61.42 & 58.86
   \\
   
& & SFT+Ours 
    & \sethlcolor{lightpink}\hl{18.07} \up{4.01} & \sethlcolor{lightpink}\hl{73.35} \up{1.94} & \sethlcolor{lightpink}\hl{64.27} \up{3.87} & \sethlcolor{lightpink}\hl{61.17} \up{3.14} & 
    \sethlcolor{lightpink}\hl{17.60} \up{1.86} & 
    \sethlcolor{lightpink}\hl{73.77} \up{1.65} & 
    \sethlcolor{lightpink}\hl{64.63} \up{3.21} & 
    \sethlcolor{lightpink}\hl{61.49} \up{2.63}  
   \\

& & DFT 
    & 14.69  & 71.90 & 60.27 & 58.81 & 16.52 & 72.65 & 61.64 & 60.04 
   \\

& & {DFT+Ours}
    & \sethlcolor{lightpink}\hl{16.33} \up{1.64} & \sethlcolor{lightpink}\hl{72.54} \up{0.64} & \sethlcolor{lightpink}\hl{62.44} \up{2.17} & \sethlcolor{lightpink}\hl{59.58} \up{0.77} & 
    \sethlcolor{lightpink}\hl{17.04} \up{0.52} & 
    \sethlcolor{lightpink}\hl{73.16} \up{0.51} & 
    \sethlcolor{lightpink}\hl{63.01} \up{1.37} & 
    \sethlcolor{lightpink}\hl{60.49} \up{0.45} 
   \\ 
   
\bottomrule
\end{tabular}
}\vspace{-0.5em}
\label{tab:benchmark_results_inat_plant}
\end{table*}

\begin{table*}[ht]
\centering
\small
\caption{Performance evaluation on iNat-Animal with more base models. The best results are in \sethlcolor{lightpink}\hl{green}.}
\vspace{-0.35em}
\resizebox{1.0\linewidth}{!}{
\begin{tabular}{l|lcccccccc}
\toprule
  \multirow{2}{*}[-0.5ex]{{Model}} 
 & \multirow{2}{*}[-0.5ex]{{Methods}}  
 & \multicolumn{4}{c}{{Base}}  
 & \multicolumn{4}{c}{{Novel}} 
 \\ 
\cmidrule(l){3-6}\cmidrule(l){7-10}
&  
& {HCA} $\uparrow$
& {POR} $\uparrow$ 
& {S-POR} $\uparrow$ 
& {TOR} $\uparrow$
& {HCA} $\uparrow$
& {POR} $\uparrow$ 
& {S-POR} $\uparrow$ 
& {TOR} $\uparrow$

\\ 
\midrule

 \multirow{2}{*}[-0ex]{Intern3.5-VL-1B}

&  SFT  
    & 2.04 & 51.13 & 37.07 & 32.32 & 1.63 & 51.33 & 36.58 & 32.44
   \\
   
& SFT+Ours 
    & \sethlcolor{lightpink}\hl{3.04} \up{1.00} & \sethlcolor{lightpink}\hl{53.86} \up{2.73} & \sethlcolor{lightpink}\hl{40.64} \up{3.57} & \sethlcolor{lightpink}\hl{35.67} \up{3.35} & \sethlcolor{lightpink}\hl{2.93} \up{1.30} & \sethlcolor{lightpink}\hl{54.17} \up{2.84} & \sethlcolor{lightpink}\hl{40.80} \up{4.22} & \sethlcolor{lightpink}\hl{36.06} \up{3.62}

   \\ \cmidrule(r){1-10}

 \multirow{2}{*}[-0ex]{LLaVA-OV-1.5-4B}

&  SFT  
    & 5.75 & 61.68 & 51.55 & 47.06 & 6.05 & 61.50 & 51.62 & 47.29
   \\
   
& SFT+Ours 
    & \sethlcolor{lightpink}\hl{6.49} \up{0.74} & \sethlcolor{lightpink}\hl{63.45} \up{1.77} & \sethlcolor{lightpink}\hl{54.20} \up{2.65} & \sethlcolor{lightpink}\hl{49.70} \up{2.64} & \sethlcolor{lightpink}\hl{6.09} \up{0.04} & \sethlcolor{lightpink}\hl{63.22} \up{1.72} & \sethlcolor{lightpink}\hl{53.89} \up{2.27} & \sethlcolor{lightpink}\hl{49.66} \up{2.37}

   \\ \cmidrule(r){1-10}

 \multirow{2}{*}[-0ex]{{Qwen2-VL-2B}}

&  SFT  
    & 6.12 & 61.88 & 51.22 & 47.18 & 5.31 & 61.77 & 51.38 & 46.77 
   \\
   
& SFT+Ours 
    & \sethlcolor{lightpink}\hl{6.42} \up{0.30} & \sethlcolor{lightpink}\hl{63.14} \up{1.26} & \sethlcolor{lightpink}\hl{54.15} \up{2.93} & \sethlcolor{lightpink}\hl{49.03} \up{1.85} & \sethlcolor{lightpink}\hl{5.94} \up{0.63} & \sethlcolor{lightpink}\hl{63.08} \up{1.31} & \sethlcolor{lightpink}\hl{54.32} \up{2.94} & \sethlcolor{lightpink}\hl{49.13} \up{2.36}

   \\

\bottomrule
\end{tabular}
}\vspace{-0.5em}
\label{tab:other_models}
\end{table*}

\begin{table*}[ht]
\centering
\caption{Performance evaluation on CUB-200 with four-level hierarchy. The best results are in \sethlcolor{lightpink}\hl{green}. }
\vspace{-0.35em}
\resizebox{1.0\linewidth}{!}{
\begin{tabular}{ll|lcccccccc}
\toprule
\multirow{2}{*}[-0.5ex]{{Dataset}} & \multirow{2}{*}[-0.5ex]{{Model}} 
 & \multirow{2}{*}[-0.5ex]{{Methods}}  
 & \multicolumn{4}{c}{{Base}} 
 & \multicolumn{4}{c}{{Novel}}  
 \\ 
\cmidrule(l){4-7} \cmidrule(l){8-11} 
&  
&
& {HCA} $\uparrow$
& {POR} $\uparrow$ 
& {S-POR} $\uparrow$ 
& {TOR} $\uparrow$
& {HCA} $\uparrow$
& {POR} $\uparrow$ 
& {S-POR} $\uparrow$ 
& {TOR} $\uparrow$
\\ 
\midrule

\multirow{2}{*}[-0ex]{{CUB-200}} & \multirow{2}{*}[-0ex]{{Qwen2.5-VL-3B}}

 & SFT  
    & 9.36 & 53.01 & 15.81 & 33.08 & 8.76 & 51.48 & 16.28 & 33.54
   \\
   
& & SFT+Ours 
    & \sethlcolor{lightpink}\hl{35.38} \up{26.02} 
    & \sethlcolor{lightpink}\hl{76.30} \up{23.29}
    & \sethlcolor{lightpink}\hl{63.38} \up{47.57}
    & \sethlcolor{lightpink}\hl{60.52} \up{27.44}
    & \sethlcolor{lightpink}\hl{32.83} \up{24.07}
    & \sethlcolor{lightpink}\hl{74.79} \up{23.31}
    & \sethlcolor{lightpink}\hl{59.06} \up{42.78}
    & \sethlcolor{lightpink}\hl{58.25} \up{24.71}
   \\

\bottomrule
\end{tabular}
}\vspace{-0.5em}
\label{tab:cub}
\end{table*}

\subsection{Variants of Dispersive Loss} 

We investigate several alternative designs of the dispersive loss by varying the geometric space in which sibling separation is measured, while keeping the same margin-based formulation. As summarized in Table~\ref{tab:ablation_disp_variants}, these variants include: (1) Variant I: hyperbolic geodesic distance $d_{\mathbb{L}}(\boldsymbol{z}_i,\boldsymbol{z}_j)$ in the Lorentz model. (2) Variant II: angular distance in the tangent space at the hyperbolic origin where $d_{\mathrm{ang}}(\mathbf{u}, \mathbf{v}) = \arccos \frac{\mathbf{u}^\top \mathbf{v}}{\|\mathbf{u}\|_2 \, \|\mathbf{v}\|_2}, 
\quad \mathbf{u}, \mathbf{v} \in \mathbb{R}^D$. (3) Variant III: parent-centered tangent-space angular separation, and (4) Variant IV: parent-relative spherical angular separation.
While these designs are geometrically motivated, they either entangle angular dispersion with radial hierarchy, rely on global reference points, or introduce distortion through repeated logarithmic mappings.

To analyze the interaction between dispersive objectives and the radial-angular decomposition of hyperbolic embeddings, 
we unify the input space for all sibling embeddings as the unit hypersphere
\begin{equation}
\small
\hat{\boldsymbol{v}} \in \mathbb{S}^{D-1} 
= \left\{ \boldsymbol{u}\in\mathbb{R}^D \;\middle|\; \|\boldsymbol{u}\|_2 = 1 \right\},
\end{equation}
where \(\hat{\boldsymbol{v}}\) is obtained via angular normalization of Euclidean visual features, 
and let \(\boldsymbol{z} \in \mathbb{H}^D\) denote the corresponding hyperbolic embeddings under the exponential map \(\exp^{\kappa}_{\mathbf{0}}\) with radial component \(\rho(\boldsymbol{z})\).

\textbf{Theorem 4.1.}
\emph{For dispersive objectives defined using hyperbolic geodesic distances or tangent-space angular measures, 
optimization generally induces gradients along the radial directions of hyperbolic embeddings, even when the inputs are spherical embeddings \(\hat{\boldsymbol{v}}_i, \hat{\boldsymbol{v}}_j \in \mathbb{S}^{D-1}\).}

Formally, let \(\hat{\boldsymbol{v}}_i, \hat{\boldsymbol{v}}_j \in \mathbb{S}^{D-1}\) be sibling embeddings, and \(\boldsymbol{z}_i, \boldsymbol{z}_j \in \mathbb{H}^D\) their hyperbolic counterparts. 
If the loss depends on \(d_{\mathbb{H}}(\boldsymbol{z}_i,\boldsymbol{z}_j)\) or on angular measures computed via \(\log_{\boldsymbol{c}}(\cdot)\) for any reference point \(\boldsymbol{c}\), 
then, except for degenerate configurations, 
\(\nabla_{\rho(\boldsymbol{z}_i)} \mathcal{L} \neq 0\) and 
\(\nabla_{\rho(\boldsymbol{z}_j)} \mathcal{L} \neq 0\).

\textbf{Theorem 4.2.}
\emph{The proposed spherical angular dispersive loss, defined on \(\hat{\boldsymbol{v}} \in \mathbb{S}^{D-1}\), does not induce gradients on the radial components of hyperbolic embeddings.}

Specifically, for any sibling embedding \(\hat{\boldsymbol{v}}_i \in \mathbb{S}^{D-1}\) with \(\boldsymbol{z}_i = \exp^{\kappa}_{\mathbf{0}}(\hat{\boldsymbol{v}}_i)\),
\begin{equation}
\small
\frac{\partial \mathcal{L}_{\mathrm{disp}}}{\partial \rho(\boldsymbol{z}_i)} = 0,
\end{equation}
ensuring that the hierarchical structure encoded by hyperbolic radii remains unchanged.

\textbf{Theorem 4.3.}
\emph{Increasing spherical angular separation between sibling embeddings on \(\mathbb{S}^{D-1}\) increases their angular separation in hyperbolic space.}

For \(\hat{\boldsymbol{v}}_i, \hat{\boldsymbol{v}}_j \in \mathbb{S}^{D-1}\) and \(\boldsymbol{z}_i = \exp^{\kappa}_{\mathbf{0}}(\hat{\boldsymbol{v}}_i), \boldsymbol{z}_j = \exp^{\kappa}_{\mathbf{0}}(\hat{\boldsymbol{v}}_j)\),
enlarging
\(\angle(\hat{\boldsymbol{v}}_i,\hat{\boldsymbol{v}}_j)\)
monotonically increases
\(\angle(\boldsymbol{z}_i,\boldsymbol{z}_j)\),
improving sibling discriminability without affecting radial ordering.

\textbf{Intuition.}
Hyperbolic hierarchy is primarily encoded by radial depth \(\rho(\boldsymbol{z})\), while sibling discrimination is governed by angular structure on \(\mathbb{S}^{D-1}\). 
Most existing dispersive variants entangle these two factors through hyperbolic distances or tangent-space mappings. 
By operating on spherical embeddings, our loss preserves radial hierarchy while enhancing angular separation, yielding stable and hierarchically consistent representations in deep taxonomies.

\section{Experiments}
\label{sec:exp}

\subsection{Experimental Setup}

\textbf{Datasets and Tasks.} 
To rigorously evaluate the effectiveness of \ours, we conduct comprehensive experiments on the iNaturalist-2021 (iNat21) dataset~\cite{inat2021}. 
Following~\cite{tan2025vision}, the dataset is divided into two taxonomies, \emph{Plant} and \emph{Animal}, comprising 4,271 and 5,388 leaf nodes, respectively, organized into six hierarchical levels. 
Unless otherwise specified, each taxonomy is evenly split into base and novel classes, and all experiments are performed under the 1-shot training setting. 
Models are trained on base classes and evaluated on both base classes (base-to-base) and novel classes (base-to-novel).

\textbf{Evaluation Metrics.} 
We assess LMMs' prediction consistency in hierarchical visual recognition task using four metrics: Hierarchical Consistent Accuracy (HCA)~\cite{protect,park2024learning}, 
Point-Overlap Ratio (POR)~\cite{por}, 
Strict Point-Overlap Ratio (S-POR)~\cite{tan2025vision}, 
and Top Overlap Ratio (TOR)~\cite{protect}. 
Detailed definitions are provided in Appendix~\ref{metrics}.

We defer additional evaluation settings and implementation details to Appendix~\ref{evaluation_setting} and Appendix~\ref{implementation_datails}, respectively.

\subsection{Main Results}

Table~\ref{tab:benchmark_results_inat_plant} compares \ours\ with widely used LMM fine-tuning methods on the Qwen2.5-VL-3B~\cite{Qwen2.5-VL} model across different taxonomies. Several key observations can be drawn. (1) Compatibility with different LMM fine-tuning methods: \ours\ consistently improves performance under both supervised fine-tuning (SFT) and dynamic fine-tuning (DFT)~\cite{wu2025generalization}. The gains in all metrics are observed across all settings, indicating that the proposed hierarchical regularization is complementary to existing LMM fine-tuning strategies rather than being tied to a specific training paradigm. (2) Robustness across taxonomies and base-to-novel generalization: Consistent improvements are achieved on both iNat-Plant and iNat-Animal, despite their distinct taxonomic structures and visual characteristics. Moreover, \ours\ improves performance on both base and novel classes in the base-to-novel evaluation, demonstrating strong generalization under distribution shift and robustness to unseen categories. (3) Balancing hierarchical consistency and discriminability: HCA, the most strict metric, requires predictions to be correct at all taxonomic levels simultaneously, and thus cannot be improved by enhancing hierarchical consistency or per-level discrimination alone. The consistent HCA gains indicate that \ours\ effectively balances these two factors, yielding coherent and discriminative representations throughout the entire taxonomy.

\begin{table}[t]
  \centering
  \small
  \caption{Ablation study on variants of dispersive loss.}
    \fontsize{7pt}{8pt}\selectfont
    \setlength{\tabcolsep}{12pt}
  \begin{tabular}{c|ll|ll}
    \toprule
    \multirow{2}{*}{Variant}  & \multicolumn{2}{c|}{Base}&  \multicolumn{2}{c}{Novel} \\
      \cmidrule(lr){2-3} \cmidrule(lr){4-5}  & HCA & $\mathrm{Acc_{leaf}}$ & HCA & $\mathrm{Acc_{leaf}}$\\
    \midrule
    I   &  15.33 & 39.78 & 16.90 & 40.62 \\
    II & 16.73 & 39.85 & 16.86 & 40.59 \\
    III & 17.92 & 40.67 & \sethlcolor{lightpink}\hl{18.23} & 42.00\\
    IV & 17.89 & 40.74 & 18.01 & 41.14 \\
    \midrule
    Ours & \sethlcolor{lightpink}\hl{18.07} & \sethlcolor{lightpink}\hl{41.37} & 17.60 & \sethlcolor{lightpink}\hl{42.52} \\
    \bottomrule
  \end{tabular}
  \label{tab:ablation_disp_variants}
\end{table}

\begin{table}[t]
  \centering
  \small
  \caption{Ablation study on key design components.
  }
    \fontsize{7pt}{8pt}\selectfont
    \setlength{\tabcolsep}{12pt}
  \begin{tabular}{ccc|cc}
    \toprule
   \multirow{1}{*}{$\mathcal{L}_\mathbf{ent}$} & \multirow{1}{*}{$\mathcal{L}_\mathbf{disp}$} & \multirow{1}{*}{Layer Index}  & Base & Novel \\
    \midrule
    \multicolumn{3}{l|}{Zero-shot} & 15.92 & 39.00  \\
    \multicolumn{3}{l|}{SFT} & 14.06 & 40.37  \\
    \midrule
    \multicolumn{5}{l}{\textit{Ablation on different regularization loss}}\\
    \checkmark & & 28 & 17.89 & 40.78 \\
    \checkmark & \checkmark & 28 & \sethlcolor{lightpink}\hl{18.07} & \sethlcolor{lightpink}\hl{41.37}  \\
    \midrule
    \multicolumn{5}{l}{\textit{Ablation on different regularization layer}}\\

    \checkmark & \checkmark & 7 & 14.55 & 40.52 \\
    \checkmark & \checkmark & 14 & 14.36 & 39.04 \\
   \checkmark & \checkmark &  21 & 11.32 & 37.63 \\
    \checkmark & \checkmark & 28 & \sethlcolor{lightpink}\hl{18.07} & \sethlcolor{lightpink}\hl{41.37}  \\
    \checkmark & \checkmark & 7,14,21,28 & 15.18 & 39.93 \\
    \checkmark & \checkmark & all layers & 12.06 & 38.81 \\
    \bottomrule
  \end{tabular}
  \label{tab:ablation_layer}
\end{table}

\subsection{Ablation Studies}

To validate the contributions of individual components in \ours, we conduct extensive ablation studies. 
Unless otherwise stated, all ablation experiments are performed on Qwen2.5-VL-3B using the iNat-Animal taxonomy.

\textbf{Evaluation on Other Models.} 
Thus far, our experiments have focused on Qwen2.5-VL-3B model. In Table~\ref{tab:other_models}, we extend our evaluation to Intern3.5-VL-1B~\cite{wang2025internvl3_5}, LLaVA-OV-1.5-4B\cite{LLaVA-OneVision-1.5}, and Qwen2-VL-2B\cite{Qwen2-vl}. Once again, \ours~consistently improves performance over the baselines across all metrics.

\textbf{Evaluation on Other Taxonomies.} 
We further examine whether \ours~generalizes to different taxonomies. 
Experiments are conducted on the CUB-200-2011 (CUB-200) dataset~\cite{cub200}, which adopts a four-level hierarchy where leaf nodes correspond to bird common names rather than scientific names~\cite{tan2025vision}. 
We use all training samples from 100 base classes and evaluate performance on both base and novel classes. 
As shown in Table~\ref{tab:cub}, \ours~consistently improves all evaluation metrics, indicating its adaptability across datasets and its effectiveness as a regularizer beyond standard fine-tuning.

\textbf{Variants of Dispersive Loss.} Table~\ref{tab:ablation_disp_variants} compares different dispersive loss variants that measure sibling separation in different geometric spaces. Variants defined in hyperbolic space or tangent spaces (I–IV) improve HCA to varying degrees, but yield limited gains on the leaf-node accuracy $\mathrm{Acc_{leaf}}$ due to interference with the radial structure that encodes taxonomy. In contrast, our spherical angular dispersive loss achieves the best consistency–discriminability trade-off by computing angular separation directly on the unit hypersphere, demonstrating the strongest improvements on $\mathrm{Acc_{leaf}}$, especially at the deepest level of the hierarchy.

\textbf{Effects of $\mathcal{L}_\mathrm{ent}$ and $\mathcal{L}_\mathrm{disp}$.} 
Table~\ref{tab:ablation_layer} shows that SFT on VQA data with taxonomic labels alone may even hurt the discrimination of similar base classes, suggesting that SFT is insufficient to induce taxonomic structure. Introducing $\mathcal{L}_{\mathrm{ent}}$ enforces hierarchical consistency and yields clear gains, while $\mathcal{L}_{\mathrm{disp}}$ further improves fine-grained separability. Combining both losses achieves the best overall performance, demonstrating their complementary effects in \ours.

\textbf{Target Regularization Layer.} 
We further investigate the effect of applying regularization at different layers. 
As shown in Table~\ref{tab:ablation_layer}, performance varies across layers, with regularizing the last layer producing the strongest results.

\begin{figure}
    \centering
    \includegraphics[width=0.45\linewidth]{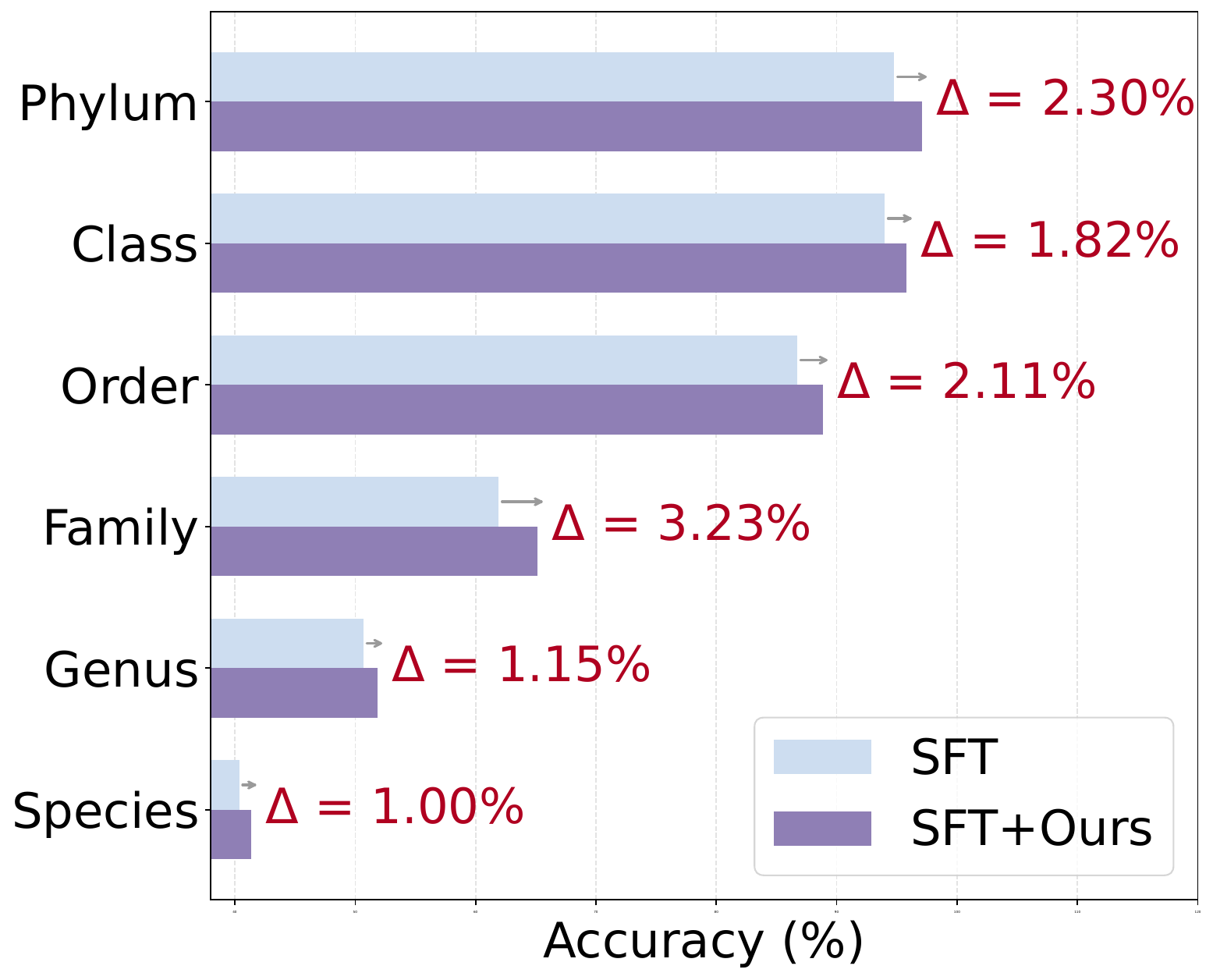}
    \hspace{0.05\linewidth} 
    \includegraphics[width=0.45\linewidth]{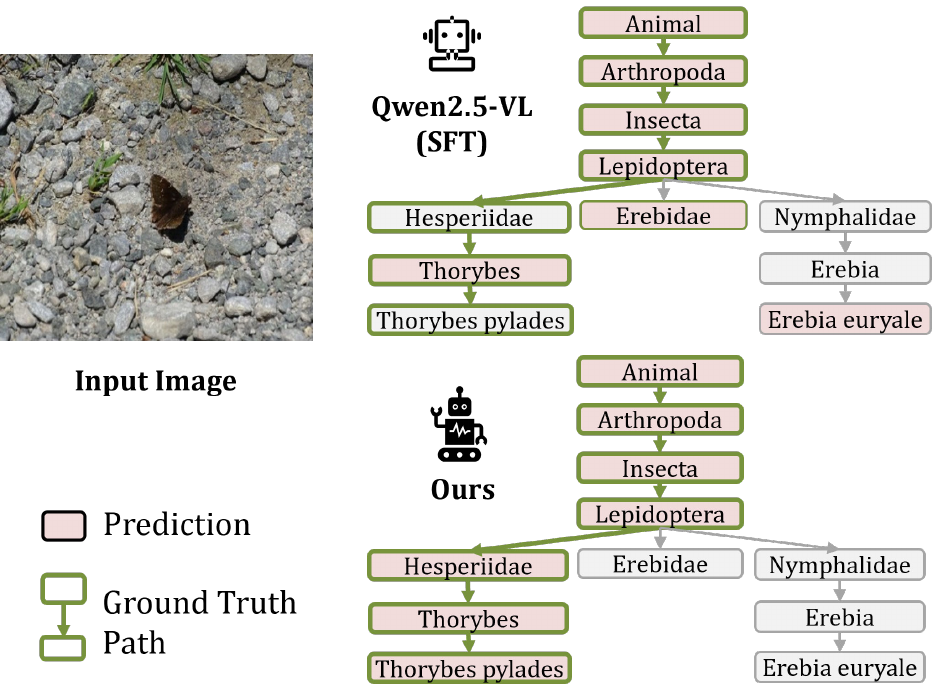} 
    \caption{Left: Per-level accuracy on iNat-Animal base classes; Right: Qualitative comparsion.}
    \label{fig:per_level_accuracy}
\end{figure}

\textbf{Per-level Accuracy.} 
Figure~\ref{fig:per_level_accuracy} (Left) shows that \ours~consistently improves accuracy across all hierarchy levels, with larger gains at coarser levels. This indicates that the improvements mainly stem from enhanced hierarchical consistency rather than purely improved fine-grained discrimination, underscoring the benefit of our approach beyond FGVR.

\textbf{Qualitative Results.}
As shown in Figure~\ref{fig:per_level_accuracy} (Right), \ours~produces consistent predictions along the taxonomy, whereas the SFT-based baseline yields inconsistent outputs, highlighting the importance of explicitly learning taxonomic structures for hierarchical visual recognition.

More ablation studies are in Appendix~\ref{ablation_param}.

\section{Conclusion}

In this work, we have proposed Hierarchical Representation Regularization (\ours), an approach that regularizes the internal representations of LMMs for learning taxonomic trees. Our experiments across various taxonomies, base models, and fine-tuning strategies demonstrate the effectiveness of \ours in improving the performance of HVR. Importantly, a key principle guiding our design is to introduce minimal or no interference with the language modeling process of the original training objective. This allows \ours compatible with any LMM and fine-tuning strategy like supervised fine-tuning, emphasizing its widespread applicability.

\section*{Acknowledgements}
This work was supported by the grants from Beijing Natural Science Foundation (L247006) and the National Natural Science Foundation of China (62525201, 62132001, 62432001).

\section*{Impact Statement}
This paper presents work whose goal is to advance the field of Machine Learning. There are many potential societal consequences of our work, none which we feel must be specifically highlighted here.





\nocite{langley00}

\bibliography{example_paper}

@article{he2026taxonomy,
  title={Taxonomy-Aware Representation Alignment for Hierarchical Visual Recognition with Large Multimodal Models},
  author={He, Hulingxiao and Tan, Zhi and Peng, Yuxin},
  journal={arXiv preprint arXiv:2603.00431},
  year={2026}
}

@article{he2026fine,
  title={Fine-r1: Make multi-modal llms excel in fine-grained visual recognition by chain-of-thought reasoning},
  author={He, Hulingxiao and Geng, Zijun and Peng, Yuxin},
  journal={arXiv preprint arXiv:2602.07605},
  year={2026}
}

@article{ye2022unsupervised,
  title={Unsupervised Cross-Media Hashing Learning via Knowledge Graph},
  author={Ye, Zhaoda and He, Xiangteng and Peng, Yuxin},
  journal={Chinese Journal of Electronics},
  volume={31},
  number={6},
  pages={1081--1091},
  year={2022},
  publisher={Wiley Online Library}
}

@article{xiao2026survey,
  title={A Survey on the Green Development of Large Models: From Resource-Efficient Architectures to Hardware-Software Co-Design},
  author={Xiao, Linhui and Cao, Guiping and Guo, Mingyue and Guan, Xianchao and Yang, Fan and Tao, Ming and Li, Xin and Peng, Yuxin and Wang, Yaowei},
  journal={Chinese Journal of Electronics},
  volume={35},
  number={5},
  pages={1--24},
  year={2026}
}

@article{peng2025survey,
  title={A survey on fine-grained multimodal large language models},
  author={Peng, Yuxin and Wang, Zishuo and Zheng, Xiangtian and Li, Geng and Yin, Sibo and He, Hulingxiao},
  year={2025},
  publisher={TechRxiv}
}

@InProceedings{Wang_2024_CVPR,
    author    = {Wang, Yuan and Li, Yali and Wang, Shengjin},
    title     = {G{\textasciicircum}3-LQ: Marrying Hyperbolic Alignment with Explicit Semantic-Geometric Modeling for 3D Visual Grounding},
    booktitle = {Proceedings of the IEEE/CVF Conference on Computer Vision and Pattern Recognition},
    month     = {June},
    year      = {2024},
    pages     = {13917-13926}
}

@book{lee2006riemannian,
	title={Riemannian manifolds: an introduction to curvature},
	author={Lee, John M},
	volume={176},
	year={2006},
	publisher={Springer Science \& Business Media}
}

@article{cannon1997hyperbolic,
  title={Hyperbolic geometry},
  author={Cannon, James W and Floyd, William J and Kenyon, Richard and Parry, Walter R and others},
  journal={Flavors of geometry},
  volume={31},
  number={59-115},
  pages={2},
  year={1997}
}

@article{ganea2018hyperbolic,
  title={Hyperbolic neural networks},
  author={Ganea, Octavian and B{\'e}cigneul, Gary and Hofmann, Thomas},
  journal={Advances in Neural Information Processing Systems},
  volume={31},
  year={2018}
}

@article{shimizu2020hyperbolic,
  title={Hyperbolic neural networks++},
  author={Shimizu, Ryohei and Mukuta, Yusuke and Harada, Tatsuya},
  journal={arXiv preprint arXiv:2006.08210},
  year={2020}
}

@inproceedings{guo2022clipped,
  title={Clipped hyperbolic classifiers are super-hyperbolic classifiers},
  author={Guo, Yunhui and Wang, Xudong and Chen, Yubei and Yu, Stella X},
  booktitle={Proceedings of the IEEE/CVF Conference on Computer Vision and Pattern Recognition},
  pages={11--20},
  year={2022}
}

@inproceedings{skopek2019mixed,
  title={Mixed-curvature Variational Autoencoders},
  author={Skopek, Ondrej and Ganea, Octavian-Eugen and B{\'e}cigneul, Gary},
  booktitle={International Conference on Learning Representations},
  year={2020}
}

@inproceedings{fu2023hyperbolic,
  title={Hyperbolic Geometric Graph Representation Learning for Hierarchy-imbalance Node Classification},
  author={Fu, Xingcheng and Wei, Yuecen and Sun, Qingyun and Yuan, Haonan and Wu, Jia and Peng, Hao and Li, Jianxin},
  booktitle={Proceedings of the ACM Web Conference 2023},
  pages={460--468},
  year={2023}
}

@inproceedings{franco2023hyperbolic_al,
  title={Hyperbolic Active Learning for Semantic Segmentation under Domain Shift},
  author={Franco, Luca and Mandica, Paolo and Kallidromitis, Konstantinos and Guillory, Devin and Li, Yu-Teng and Darrell, Trevor and Galasso, Fabio},
  booktitle={International Conference on Machine Learning},
  year={2024},
  organization={PMLR}
}

@inproceedings{ramasinghe2024accept,
  title={Accept the modality gap: An exploration in the hyperbolic space},
  author={Ramasinghe, Sameera and Shevchenko, Violetta and Avraham, Gil and Thalaiyasingam, Ajanthan},
  booktitle={Proceedings of the IEEE/CVF Conference on Computer Vision and Pattern Recognition},
  pages={27263--27272},
  year={2024}
}

@inproceedings{mandica2024hyperbolic,
  title={Hyperbolic learning with multimodal large language models},
  author={Mandica, Paolo and Franco, Luca and Kallidromitis, Konstantinos and Petryk, Suzanne and Galasso, Fabio},
  booktitle={European Conference on Computer Vision},
  pages={382--398},
  year={2024},
  organization={Springer}
}

@article{wang2024cliploss,
  title={Cliploss and norm-based data selection methods for multimodal contrastive learning},
  author={Wang, Yiping and Chen, Yifang and Yan, Wendan and Fang, Alex and Zhou, Wenjing and Jamieson, Kevin and Du, Simon S},
  journal={Advances in Neural Information Processing Systems},
  volume={37},
  pages={15028--15069},
  year={2024}
}

@inproceedings{desai2023hyperbolic,
  title={Hyperbolic image-text representations},
  author={Desai, Karan and Nickel, Maximilian and Rajpurohit, Tanmay and Johnson, Justin and Vedantam, Shanmukha Ramakrishna},
  booktitle={International Conference on Machine Learning},
  pages={7694--7731},
  year={2023},
  organization={PMLR}
}

@inproceedings{palcompositional,
  title={Compositional Entailment Learning for Hyperbolic Vision-Language Models},
  author={Pal, Avik and van Spengler, Max and di Melendugno, Guido Maria D'Amely and Flaborea, Alessandro and Galasso, Fabio and Mettes, Pascal},
  booktitle={The Thirteenth International Conference on Learning Representations},
  year={2025},
}

@article{hong2023curved,
  title={Curved geometric networks for visual anomaly recognition},
  author={Hong, Jie and Fang, Pengfei and Li, Weihao and Han, Junlin and Petersson, Lars and Harandi, Mehrtash},
  journal={IEEE Transactions on Neural Networks and Learning Systems},
  year={2023},
  publisher={IEEE}
}

@inproceedings{long2020searching,
  title={Searching for actions on the hyperbole},
  author={Long, Teng and Mettes, Pascal and Shen, Heng Tao and Snoek, Cees GM},
  booktitle={Proceedings of the IEEE/CVF Conference on Computer Vision and Pattern Recognition},
  pages={1141--1150},
  year={2020}
}

@inproceedings{radford2021learning,
  title={Learning transferable visual models from natural language supervision},
  author={Radford, Alec and Kim, Jong Wook and Hallacy, Chris and Ramesh, Aditya and Goh, Gabriel and Agarwal, Sandhini and Sastry, Girish and Askell, Amanda and Mishkin, Pamela and Clark, Jack and others},
  booktitle={International conference on machine learning},
  pages={8748--8763},
  year={2021},
  organization={PMLR}
}

@inproceedings{stevens2024bioclip,
  title={Bioclip: A vision foundation model for the tree of life},
  author={Stevens, Samuel and Wu, Jiaman and Thompson, Matthew J and Campolongo, Elizabeth G and Song, Chan Hee and Carlyn, David Edward and Dong, Li and Dahdul, Wasila M and Stewart, Charles and Berger-Wolf, Tanya and others},
  booktitle={Proceedings of the IEEE/CVF conference on computer vision and pattern recognition},
  pages={19412--19424},
  year={2024}
}

@inproceedings{malik2025hyperdefender,
  title={Hyperdefender: A robust framework for hyperbolic gnns},
  author={Malik, Nikita and Gupta, Rahul and Kumar, Sandeep},
  booktitle={Proceedings of the AAAI Conference on Artificial Intelligence},
  volume={39},
  number={18},
  pages={19396--19404},
  year={2025}
}

@inproceedings{li2025hyperbolic,
  title={Hyperbolic-constraint Point Cloud Reconstruction from Single RGB-D Images},
  author={Li, Wenrui and Yang, Zhe and Han, Wei and Man, Hengyu and Wang, Xingtao and Fan, Xiaopeng},
  booktitle={Proceedings of the AAAI Conference on Artificial Intelligence},
  volume={39},
  number={5},
  pages={4959--4967},
  year={2025}
}

@article{he2025helm,
  title={HELM: Hyperbolic Large Language Models via Mixture-of-Curvature Experts},
  author={He, Neil and Anand, Rishabh and Madhu, Hiren and Maatouk, Ali and Krishnaswamy, Smita and Tassiulas, Leandros and Yang, Menglin and Ying, Rex},
  journal={arXiv preprint arXiv:2505.24722},
  year={2025}
}

@inproceedings{he2025lorentzian,
  title={Lorentzian residual neural networks},
  author={He, Neil and Yang, Menglin and Ying, Rex},
  booktitle={Proceedings of the 31st ACM SIGKDD Conference on Knowledge Discovery and Data Mining V. 1},
  pages={436--447},
  year={2025}
}

@inproceedings{fu2024hyperbolic,
  title={Hyperbolic geometric latent diffusion model for graph generation},
  author={Fu, Xingcheng and Gao, Yisen and Wei, Yuecen and Sun, Qingyun and Peng, Hao and Li, Jianxin and Li, Xianxian},
  booktitle={International Conference on Machine Learning (ICML)},
  year={2024}
}

@article{fan2025curvature,
  title={Curvature Learning for Generalization of Hyperbolic Neural Networks: X. Fan et al.},
  author={Fan, Xiaomeng and Wu, Yuwei and Gao, Zhi and Harandi, Mehrtash and Jia, Yunde},
  journal={International Journal of Computer Vision},
  volume={133},
  number={12},
  pages={8489--8525},
  year={2025},
  publisher={Springer}
}

@article{li2025geometry,
  title={Geometry-aware Distance Measure for Diverse Hierarchical Structures in Hyperbolic Spaces},
  author={Li, Pengxiang and Wu, Yuwei and Gao, Zhi and Fan, Xiaomeng and Wu, Wei and Lu, Zhipeng and Jia, Yunde and Harandi, Mehrtash},
  journal={arXiv preprint arXiv:2506.18533},
  year={2025}
}

@inproceedings{gao2021curvature,
  title={Curvature generation in curved spaces for few-shot learning},
  author={Gao, Zhi and Wu, Yuwei and Jia, Yunde and Harandi, Mehrtash},
  booktitle={Proceedings of the IEEE/CVF International Conference on Computer Vision},
  pages={8691--8700},
  year={2021}
}

@inproceedings{gao2023exploring,
  title={Exploring Data Geometry for Continual Learning},
  author={Gao, Zhi and Xu, Chen and Li, Feng and Jia, Yunde and Harandi, Mehrtash and Wu, Yuwei},
  booktitle={Proceedings of the IEEE/CVF Conference on Computer Vision and Pattern Recognition},
  pages={24325--24334},
  year={2023}
}

@article{yu2025hyperbolic,
  title={Hyperbolic Dual Feature Augmentation for Open-Environment},
  author={Yu, Peilin and Wu, Yuwei and Gao, Zhi and Fan, Xiaomeng and Yang, Shuo and Jia, Yunde},
  journal={arXiv preprint arXiv:2506.08906},
  year={2025}
}

@article{khosla2020supervised,
  title={Supervised contrastive learning},
  author={Khosla, Prannay and Teterwak, Piotr and Wang, Chen and Sarna, Aaron and Tian, Yonglong and Isola, Phillip and Maschinot, Aaron and Liu, Ce and Krishnan, Dilip},
  journal={Advances in neural information processing systems},
  volume={33},
  pages={18661--18673},
  year={2020}
}

@inproceedings{mu2022slip,
  title={Slip: Self-supervision meets language-image pre-training},
  author={Mu, Norman and Kirillov, Alexander and Wagner, David and Xie, Saining},
  booktitle={European conference on computer vision},
  pages={529--544},
  year={2022},
  organization={Springer}
}

@article{repa,
  title={Representation alignment for generation: Training diffusion transformers is easier than you think},
  author={Yu, Sihyun and Kwak, Sangkyung and Jang, Huiwon and Jeong, Jongheon and Huang, Jonathan and Shin, Jinwoo and Xie, Saining},
  journal={arXiv preprint arXiv:2410.06940},
  year={2024}
}

@article{chen2025sara,
  title={{SARA}: Structural and Adversarial Representation Alignment for Training-efficient Diffusion Models},
  author={Chen, Hesen and Wang, Junyan and Tan, Zhiyu and Li, Hao},
  journal={arXiv preprint arXiv:2503.08253},
  year={2025}
}

@article{softrepa,
  title={Aligning Text to Image in Diffusion Models is Easier Than You Think},
  author={Lee, Jaa-Yeon and Cha, Byunghee and Kim, Jeongsol and Ye, Jong Chul},
  journal={arXiv preprint arXiv:2503.08250},
  year={2025}
}

@article{viral,
  title={Visual Representation Alignment for Multimodal Large Language Models},
  author={Yoon, Heeji and Jung, Jaewoo and Kim, Junwan and Choi, Hyungyu and Shin, Heeseong and Lim, Sangbeom and An, Honggyu and Kim, Chaehyun and Han, Jisang and Kim, Donghyun and others},
  journal={arXiv preprint arXiv:2509.07979},
  year={2025}
}

@article{jarvis,
  title={Seeing Beyond Words: Self-Supervised Visual Learning for Multimodal Large Language Models},
  author={Caffagni, Davide and Sarto, Sara and Cornia, Marcella and Baraldi, Lorenzo and Dovesi, Pier Luigi and Roohi, Shaghayegh and Granroth-Wilding, Mark and Cucchiara, Rita},
  journal={arXiv preprint arXiv:2512.15885},
  year={2025}
}

@article{wei2025modality,
  title={Modality Alignment across Trees on Heterogeneous Hyperbolic Manifolds},
  author={Wei, Wu and Fan, Xiaomeng and Wu, Yuwei and Gao, Zhi and Li, Pengxiang and Jia, Yunde and Harandi, Mehrtash},
  journal={arXiv preprint arXiv:2510.27391},
  year={2025}
}

@article{deepperception,
  title={KARL: Knowledge-Aware Reasoning and Reinforcement Learning for Knowledge-Intensive Visual Grounding},
  author={Ma, Xinyu and Ding, Ziyang and Luo, Zhicong and Chen, Chi and Guo, Zonghao and Wong, Derek F and Zhao, Zhen and Feng, Xiaoyi and Sun, Maosong},
  journal={arXiv preprint arXiv:2503.12797},
  year={2026}
}

@article{finedefics,
  title={Analyzing and Boosting the Power of Fine-Grained Visual Recognition for Multi-modal Large Language Models},
  author={He, Hulingxiao and Li, Geng and Geng, Zijun and Xu, Jinglin and Peng, Yuxin},
  journal={arXiv preprint arXiv:2501.15140},
  year={2025}
}

@inproceedings{hcast2025park,
title={Visually Consistent Hierarchical Image Classification},
author={Seulki Park and Youren Zhang and Stella X. Yu and Sara Beery and Jonathan Huang},
booktitle={The Thirteenth International Conference on Learning Representations},
year={2025},
}

@article{silla2011survey,
  title={A survey of hierarchical classification across different application domains},
  author={Silla, Carlos N and Freitas, Alex A},
  journal={Data mining and knowledge discovery},
  year={2011},
  publisher={Springer}
}

@article{tan2025vision,
  title={The LLM Bottleneck: Why Open-Source Vision LLMs Struggle with Hierarchical Visual Recognition},
  author={Tan, Yuwen and Qing, Yuan and Gong, Boqing},
  journal={arXiv preprint arXiv:2505.24840},
  year={2026}
}

@article{Qwen2.5-VL,
  title={Qwen2.5-VL Technical Report},
  author={Bai, Shuai and Chen, Keqin and Liu, Xuejing and Wang, Jialin and Ge, Wenbin and Song, Sibo and Dang, Kai and Wang, Peng and Wang, Shijie and Tang, Jun and Zhong, Humen and Zhu, Yuanzhi and Yang, Mingkun and Li, Zhaohai and Wan, Jianqiang and Wang, Pengfei and Ding, Wei and Fu, Zheren and Xu, Yiheng and Ye, Jiabo and Zhang, Xi and Xie, Tianbao and Cheng, Zesen and Zhang, Hang and Yang, Zhibo and Xu, Haiyang and Lin, Junyang},
  journal={arXiv preprint arXiv:2502.13923},
  year={2025}
}

@article{cub200, 
title={The Caltech-UCSD Birds-200-2011 Dataset}, 
publisher={California Institute of Technology}, 
author={Wah, Catherine and Branson, Steve and Welinder, Peter and Perona, Pietro and Belongie, Serge}, 
year={2011}, 
month={Jul} }

@inproceedings{inat2021,
  title={Benchmarking Representation Learning for Natural World Image Collections},
  author={Van Horn, Grant and Cole, Elijah and Beery, Sara and Wilber, Kimberly and Belongie, Serge and Mac Aodha, Oisin},
  booktitle={Computer Vision and Pattern Recognition},
  year={2021}
}

@inproceedings{parkvisually,
  title={Visually Consistent Hierarchical Image Classification},
  author={Park, Seulki and Zhang, Youren and Stella, X Yu and Beery, Sara and Huang, Jonathan},
  booktitle={The Thirteenth International Conference on Learning Representations},
year = {2025}
}

@inproceedings{siglip,
  title={Sigmoid loss for language image pre-training},
  author={Zhai, Xiaohua and Mustafa, Basil and Kolesnikov, Alexander and Beyer, Lucas},
  booktitle={Proceedings of the IEEE/CVF international conference on computer vision},
  pages={11975--11986},
  year={2023}
}

@inproceedings{
vlm_img_bad,
title={Why are Visually-Grounded Language Models Bad at Image Classification?},
author={Yuhui Zhang and Alyssa Unell and Xiaohan Wang and Dhruba Ghosh and Yuchang Su and Ludwig Schmidt and Serena Yeung-Levy},
booktitle={The Thirty-eighth Annual Conference on Neural Information Processing Systems},
year={2024},
url={https://openreview.net/forum?id=MwmmBg1VYg}
}

@inproceedings{clip,
  title={Learning transferable visual models from natural language supervision},
  author={Radford, Alec and Kim, Jong Wook and Hallacy, Chris and Ramesh, Aditya and Goh, Gabriel and Agarwal, Sandhini and Sastry, Girish and Askell, Amanda and Mishkin, Pamela and Clark, Jack and others},
  booktitle={International conference on machine learning},
  pages={8748--8763},
  year={2021},
  organization={PmLR}
}

@inproceedings{protect,
  title={Protect: Prompt tuning for taxonomic open set classification},
  author={Wu, Tz-Ying and Ho, Chih-Hui and Vasconcelos, Nuno},
  booktitle={Proceedings of the IEEE/CVF Conference on Computer Vision and Pattern Recognition},
  pages={16531--16540},
  year={2024}
}

@inproceedings{por,
  title={Exploring hierarchical graph representation for large-scale zero-shot image classification},
  author={Yi, Kai and Shen, Xiaoqian and Gou, Yunhao and Elhoseiny, Mohamed},
  booktitle={European Conference on Computer Vision},
  pages={116--132},
  year={2022},
  organization={Springer}
}

@article{liu2024revisiting,
  title={Revisiting mllms: An in-depth analysis of image classification abilities},
  author={Liu, Huan and Xiao, Lingyu and Liu, Jiangjiang and Li, Xiaofan and Feng, Ze and Yang, Sen and Wang, Jingdong},
  journal={arXiv preprint arXiv:2412.16418},
  year={2024}
}

@article{he2025analyzing,
  title={Analyzing and Boosting the Power of Fine-Grained Visual Recognition for Multi-modal Large Language Models},
  author={He, Hulingxiao and Li, Geng and Geng, Zijun and Xu, Jinglin and Peng, Yuxin},
  journal={arXiv preprint arXiv:2501.15140},
  year={2025}
}

@article{pal2024compositional,
  title={Compositional Entailment Learning for Hyperbolic Vision-Language Models},
  author={Pal, Avik and van Spengler, Max and di Melendugno, Guido Maria D'Amely and Flaborea, Alessandro and Galasso, Fabio and Mettes, Pascal},
  journal={arXiv preprint arXiv:2410.06912},
  year={2024}
}

@inproceedings{park2024learning,
  title={Visually consistent hierarchical image classification},
  author={Park, Seulki and Zhang, Youren and Yu, Stella and Beery, Sara and Huang, Jonathan},
  booktitle={International Conference on Learning Representations},
  volume={2025},
  pages={20808--20830},
  year={2025}
}

@article{zeng2024learning,
  title={Learning Structured Representations by Embedding Class Hierarchy with Fast Optimal Transport},
  author={Zeng, Siqi and Du, Sixian and Yamada, Makoto and Zhao, Han},
  journal={arXiv preprint arXiv:2410.03052},
  year={2024}
}

@article{he2024language,
  title={Language models as hierarchy encoders},
  author={He, Yuan and Yuan, Moy and Chen, Jiaoyan and Horrocks, Ian},
  journal={Advances in Neural Information Processing Systems},
  volume={37},
  pages={14690--14711},
  year={2024}
}

@article{sinha2024learning,
  title={Learning Structured Representations with Hyperbolic Embeddings},
  author={Sinha, Aditya and Zeng, Siqi and Yamada, Makoto and Zhao, Han},
  journal={Advances in Neural Information Processing Systems},
  volume={37},
  pages={91220--91259},
  year={2024}
}

@article{geigle2024african,
  title={African or european swallow? benchmarking large vision-language models for fine-grained object classification},
  author={Geigle, Gregor and Timofte, Radu and Glava{\v{s}}, Goran},
  journal={arXiv preprint arXiv:2406.14496},
  year={2024}
}

@inproceedings{zhou2025novel,
  title={A Novel Negative Sample Generation Method for Contrastive Learning in Hierarchical Text Classification},
  author={Zhou, Juncheng and Zhang, Lijuan and He, Yachen and Fan, Rongli and Zhang, Lei and Wan, Jian},
  booktitle={Proceedings of the 31st International Conference on Computational Linguistics},
  pages={5645--5655},
  year={2025}
}

@article{wang2022incorporating,
  title={Incorporating hierarchy into text encoder: a contrastive learning approach for hierarchical text classification},
  author={Wang, Zihan and Wang, Peiyi and Huang, Lianzhe and Sun, Xin and Wang, Houfeng},
  journal={arXiv preprint arXiv:2203.03825},
  year={2022}
}

@article{geng2023hiclip,
  title={HiCLIP: Contrastive language-image pretraining with hierarchy-aware attention},
  author={Geng, Shijie and Yuan, Jianbo and Tian, Yu and Chen, Yuxiao and Zhang, Yongfeng},
  journal={arXiv preprint arXiv:2303.02995},
  year={2023}
}

@article{xia2023hgclip,
  title={HGCLIP: exploring vision-language models with graph representations for hierarchical understanding},
  author={Xia, Peng and Yu, Xingtong and Hu, Ming and Ju, Lie and Wang, Zhiyong and Duan, Peibo and Ge, Zongyuan},
  journal={arXiv preprint arXiv:2311.14064},
  year={2023}
}

@article{yu2025benchmarking,
  title={Benchmarking Large Vision-Language Models on Fine-Grained Image Tasks: A Comprehensive Evaluation},
  author={Yu, Hong-Tao and Wei, Xiu-Shen and Peng, Yuxin and Belongie, Serge},
  journal={arXiv preprint arXiv:2504.14988},
  year={2025}
}

@article{conti2025large,
  title={On Large Multimodal Models as Open-World Image Classifiers},
  author={Conti, Alessandro and Mancini, Massimiliano and Fini, Enrico and Wang, Yiming and Rota, Paolo and Ricci, Elisa},
  journal={arXiv preprint arXiv:2503.21851},
  year={2025}
}

@inproceedings{novack2023chils,
  title={Chils: Zero-shot image classification with hierarchical label sets},
  author={Novack, Zachary and McAuley, Julian and Lipton, Zachary Chase and Garg, Saurabh},
  booktitle={International Conference on Machine Learning},
  pages={26342--26362},
  year={2023},
  organization={PMLR}
}

@inproceedings{chen2022label,
  title={Label relation graphs enhanced hierarchical residual network for hierarchical multi-granularity classification},
  author={Chen, Jingzhou and Wang, Peng and Liu, Jian and Qian, Yuntao},
  booktitle={Proceedings of the IEEE/CVF Conference on Computer Vision and Pattern Recognition},
  pages={4858--4867},
  year={2022}
}

@article{kosmopoulos2015evaluation,
  title={Evaluation measures for hierarchical classification: a unified view and novel approaches},
  author={Kosmopoulos, Aris and Partalas, Ioannis and Gaussier, Eric and Paliouras, Georgios and Androutsopoulos, Ion},
  journal={Data Mining and Knowledge Discovery},
  volume={29},
  pages={820--865},
  year={2015},
  publisher={Springer}
}

@inproceedings{zhou2020hierarchy,
  title={Hierarchy-aware global model for hierarchical text classification},
  author={Zhou, Jie and Ma, Chunping and Long, Dingkun and Xu, Guangwei and Ding, Ning and Zhang, Haoyu and Xie, Pengjun and Liu, Gongshen},
  booktitle={Proceedings of the 58th annual meeting of the association for computational linguistics},
  pages={1106--1117},
  year={2020}
}

@article{snaebjarnarson2025taxonomy,
  title={Taxonomy-Aware Evaluation of Vision-Language Models},
  author={Sn{\ae}bjarnarson, V{\'e}steinn and Du, Kevin and Stoehr, Niklas and Belongie, Serge and Cotterell, Ryan and Lang, Nico and Frank, Stella},
  journal={arXiv preprint arXiv:2504.05457},
  year={2025}
}

@article{Qwen2-vl,
  title={Qwen2-vl: Enhancing vision-language model's perception of the world at any resolution},
  author={Wang, Peng and Bai, Shuai and Tan, Sinan and Wang, Shijie and Fan, Zhihao and Bai, Jinze and Chen, Keqin and Liu, Xuejing and Wang, Jialin and Ge, Wenbin and others},
  journal={arXiv preprint arXiv:2409.12191},
  year={2024}
}

@inproceedings{zhang2025cross,
  title={Cross-modal information flow in multimodal large language models},
  author={Zhang, Zhi and Yadav, Srishti and Han, Fengze and Shutova, Ekaterina},
  booktitle={Proceedings of the Computer Vision and Pattern Recognition Conference},
  pages={19781--19791},
  year={2025}
}

@article{wu2025generalization,
  title={On the generalization of sft: A reinforcement learning perspective with reward rectification},
  author={Wu, Yongliang and Zhou, Yizhou and Ziheng, Zhou and Peng, Yingzhe and Ye, Xinyu and Hu, Xinting and Zhu, Wenbo and Qi, Lu and Yang, Ming-Hsuan and Yang, Xu},
  journal={arXiv preprint arXiv:2508.05629},
  year={2025}
}

@inproceedings{nickel2018learning,
  title={Learning continuous hierarchies in the lorentz model of hyperbolic geometry},
  author={Nickel, Maximillian and Kiela, Douwe},
  booktitle={International conference on machine learning},
  pages={3779--3788},
  year={2018},
  organization={PMLR}
}

@article{wang2025internvl3_5,
  title={InternVL3.5: Advancing Open-Source Multimodal Models in Versatility, Reasoning, and Efficiency},
  author={Wang, Weiyun and Gao, Zhangwei and Gu, Lixin and Pu, Hengjun and Cui, Long and Wei, Xingguang and Liu, Zhaoyang and Jing, Linglin and Ye, Shenglong and Shao, Jie and others},
  journal={arXiv preprint arXiv:2508.18265},
  year={2025}
}

@inproceedings{LLaVA-OneVision-1.5,
  title={LLaVA-OneVision-1.5: Fully Open Framework for Democratized Multimodal Training},
  author={An, Xiang and Xie, Yin and Yang, Kaicheng and Zhang, Wenkang and Zhao, Xiuwei and Cheng, Zheng and Wang, Yirui and Xu, Songcen and Chen, Changrui and Wu, Chunsheng and Tan, Huajie and Li, Chunyuan and Yang, Jing and Yu, Jie and Wang, Xiyao and Qin, Bin and Wang, Yumeng and Yan, Zizhen and Feng, Ziyong and Liu, Ziwei and Li, Bo and Deng, Jiankang},
  booktitle={arxiv},  
  year={2025}
 }
\bibliographystyle{icml2026}

\newpage
\appendix
\onecolumn
\section{Theoretical Analysis of Dispersive Loss}
\label{proof}

In this section, we provide a rigorous theoretical analysis of different variants of the dispersive loss under hyperbolic geometry.
We focus on how various dispersive objectives interact with the radial--angular decomposition of hyperbolic embeddings,
and formally justify why the proposed spherical angular formulation uniquely preserves radial hierarchy while improving sibling-level discrimination.


\paragraph{Notation.}
Let $\hat{\boldsymbol{v}} \in \mathbb{S}^{D-1}$ denote a unit-norm spherical embedding,
\begin{equation}
\small
\mathbb{S}^{D-1}
=
\left\{
\boldsymbol{u}\in\mathbb{R}^D \;\middle|\; \|\boldsymbol{u}\|_2 = 1
\right\}.
\end{equation}
Its corresponding hyperbolic embedding is obtained via the exponential map at the origin,
\begin{equation}
\small
\boldsymbol{z}
=
\exp^{\kappa}_{\mathbf{0}}(\boldsymbol{v})
\in \mathbb{H}^D,
\end{equation}
where $\boldsymbol{v} \in \mathbb{R}^D$ is a Euclidean vector with direction $\hat{\boldsymbol{v}} = \boldsymbol{v}/\|\boldsymbol{v}\|$.
We denote by $\rho(\boldsymbol{z}) = \|\boldsymbol{v}\|$ the radial component of $\boldsymbol{z}$, which encodes hierarchical depth in hyperbolic space.

Throughout this section, we consider sibling embeddings
$\hat{\boldsymbol{v}}_i, \hat{\boldsymbol{v}}_j \in \mathbb{S}^{D-1}$
and their hyperbolic counterparts
$\boldsymbol{z}_i, \boldsymbol{z}_j \in \mathbb{H}^D$.


\textbf{Theorem 4.1 (Radial Entanglement of Existing Dispersive Variants).}
\emph{
For dispersive objectives defined using hyperbolic geodesic distances
$d_{\mathbb{H}}(\boldsymbol{z}_i,\boldsymbol{z}_j)$
or angular measures computed via logarithmic maps
$\log_{\boldsymbol{c}}(\cdot)$
at any reference point $\boldsymbol{c}$,
optimization generally induces non-zero gradients on the radial components
$\rho(\boldsymbol{z}_i)$ and $\rho(\boldsymbol{z}_j)$,
even when the inputs are spherical embeddings
$\hat{\boldsymbol{v}}_i, \hat{\boldsymbol{v}}_j \in \mathbb{S}^{D-1}$.
}

\paragraph{Proof.}
We analyze the gradient of representative dispersive objectives with respect to the radial component $\rho(\boldsymbol{z}_i)$.

\emph{Case (i): Hyperbolic geodesic distance.}
The hyperbolic distance between $\boldsymbol{z}_i$ and $\boldsymbol{z}_j$ satisfies
\begin{equation}
\small
d_{\mathbb{H}}(\boldsymbol{z}_i,\boldsymbol{z}_j)
=
\frac{1}{\sqrt{\kappa}}
\operatorname{arccosh}
\!\left(
\cosh(\sqrt{\kappa}\rho_i)\cosh(\sqrt{\kappa}\rho_j)
-
\sinh(\sqrt{\kappa}\rho_i)\sinh(\sqrt{\kappa}\rho_j)
\langle \hat{\boldsymbol{v}}_i,\hat{\boldsymbol{v}}_j\rangle
\right),
\end{equation}
where $\rho_i=\rho(\boldsymbol{z}_i)$.

Taking the partial derivative with respect to $\rho_i$ yields
\begin{align}
\frac{\partial d_{\mathbb{H}}}{\partial \rho_i}
&=
\frac{
\sinh(\sqrt{\kappa}\rho_i)\cosh(\sqrt{\kappa}\rho_j)
-
\cosh(\sqrt{\kappa}\rho_i)\sinh(\sqrt{\kappa}\rho_j)
\langle \hat{\boldsymbol{v}}_i,\hat{\boldsymbol{v}}_j\rangle
}{
\sqrt{
\left(
\cosh(\sqrt{\kappa}\rho_i)\cosh(\sqrt{\kappa}\rho_j)
-
\sinh(\sqrt{\kappa}\rho_i)\sinh(\sqrt{\kappa}\rho_j)
\langle \hat{\boldsymbol{v}}_i,\hat{\boldsymbol{v}}_j\rangle
\right)^2
-
1
}
}.
\end{align}
Except for degenerate configurations (e.g., identical directions and radii),
this derivative is non-zero.
Hence, any dispersive loss depending on $d_{\mathbb{H}}$ induces gradients along radial directions.

\emph{Case (ii): Tangent-space angular objectives.}
Consider an angular dispersive term defined as
\begin{equation}
\small
\delta
=
\angle\!\left(
\log_{\boldsymbol{c}}(\boldsymbol{z}_i),
\log_{\boldsymbol{c}}(\boldsymbol{z}_j)
\right),
\end{equation}
where $\boldsymbol{c}$ is either the hyperbolic origin or a parent node.

The logarithmic map admits the form
\begin{equation}
\small
\log_{\boldsymbol{c}}(\boldsymbol{z}_i)
=
\alpha_i
\big(
\boldsymbol{z}_i
+
\kappa \langle \boldsymbol{c},\boldsymbol{z}_i\rangle \boldsymbol{c}
\big),
\end{equation}
where the scalar coefficient
\begin{equation}
\small
\alpha_i
=
\frac{
\operatorname{arccosh}(-\kappa \langle \boldsymbol{c},\boldsymbol{z}_i\rangle)
}{
\sqrt{
(\kappa \langle \boldsymbol{c},\boldsymbol{z}_i\rangle)^2 - 1
}
}
\end{equation}
depends explicitly on the Lorentz inner product
$\langle \boldsymbol{c},\boldsymbol{z}_i\rangle$,
which in turn depends on $\rho(\boldsymbol{z}_i)$.

Consequently, both the direction and magnitude of
$\log_{\boldsymbol{c}}(\boldsymbol{z}_i)$
vary with $\rho(\boldsymbol{z}_i)$,
and by the chain rule,
\begin{equation}
\small
\frac{\partial \delta}{\partial \rho(\boldsymbol{z}_i)} \neq 0
\end{equation}
in general.

Combining the above cases proves the claim.
\hfill $\square$


\textbf{Theorem 4.2 (Radial Invariance of Spherical Angular Dispersive Loss).}
\emph{
The proposed spherical angular dispersive loss, defined on
$\hat{\boldsymbol{v}} \in \mathbb{S}^{D-1}$,
induces zero gradient on the radial components of the corresponding hyperbolic embeddings.
}

\paragraph{Proof.}
The proposed dispersive loss is defined as
\begin{equation}
\small
\mathcal{L}_{\mathrm{disp}}
=
\mathcal{L}
\big(
\angle(\hat{\boldsymbol{v}}_i,\hat{\boldsymbol{v}}_j)
\big),
\qquad
\hat{\boldsymbol{v}}_i, \hat{\boldsymbol{v}}_j \in \mathbb{S}^{D-1}.
\end{equation}
By construction, $\hat{\boldsymbol{v}}_i$ depends only on the direction of $\boldsymbol{v}_i$
and is invariant to its norm.
Therefore,
\begin{equation}
\small
\frac{\partial \mathcal{L}_{\mathrm{disp}}}{\partial \|\boldsymbol{v}_i\|} = 0.
\end{equation}

Since the hyperbolic radius satisfies $\rho(\boldsymbol{z}_i)=\|\boldsymbol{v}_i\|$,
the chain rule gives
\begin{equation}
\small
\frac{\partial \mathcal{L}_{\mathrm{disp}}}{\partial \rho(\boldsymbol{z}_i)}
=
\frac{\partial \mathcal{L}_{\mathrm{disp}}}{\partial \|\boldsymbol{v}_i\|}
\cdot
\frac{\partial \|\boldsymbol{v}_i\|}{\partial \rho(\boldsymbol{z}_i)}
=
0.
\end{equation}
Hence, the proposed loss does not perturb hyperbolic radial hierarchy.
\hfill $\square$


\textbf{Theorem 4.3 (Angular Consistency Between Sphere and Hyperbolic Space).}
\emph{
Increasing spherical angular separation between sibling embeddings
$\hat{\boldsymbol{v}}_i, \hat{\boldsymbol{v}}_j \in \mathbb{S}^{D-1}$
monotonically increases their angular separation in hyperbolic space.
}

\paragraph{Proof.}
Let
$\boldsymbol{z}_i = \exp^{\kappa}_{\mathbf{0}}(\boldsymbol{v}_i)$ and
$\boldsymbol{z}_j = \exp^{\kappa}_{\mathbf{0}}(\boldsymbol{v}_j)$
with fixed radii $\rho_i$ and $\rho_j$.
Their Lorentz inner product satisfies
\begin{equation}
\small
-\kappa \langle \boldsymbol{z}_i,\boldsymbol{z}_j\rangle
=
\cosh(\sqrt{\kappa}\rho_i)\cosh(\sqrt{\kappa}\rho_j)
-
\sinh(\sqrt{\kappa}\rho_i)\sinh(\sqrt{\kappa}\rho_j)
\langle \hat{\boldsymbol{v}}_i,\hat{\boldsymbol{v}}_j\rangle.
\end{equation}

For fixed $\rho_i$ and $\rho_j$, the right-hand side is a strictly decreasing affine function of
$\langle \hat{\boldsymbol{v}}_i,\hat{\boldsymbol{v}}_j\rangle$.
Since
\begin{equation}
\small
\angle(\hat{\boldsymbol{v}}_i,\hat{\boldsymbol{v}}_j)
=
\arccos
\big(
\langle \hat{\boldsymbol{v}}_i,\hat{\boldsymbol{v}}_j\rangle
\big),
\end{equation}
increasing spherical angular separation strictly increases the hyperbolic angular separation between
$\boldsymbol{z}_i$ and $\boldsymbol{z}_j$.

Thus, optimizing spherical angular dispersion directly enhances sibling discriminability in hyperbolic space without altering radial ordering.
\hfill $\square$

\section{More Experimental Details and Results}

\subsection{Evaluation Metrics}
\label{metrics}
To comprehensively evaluate model performance, we focus on the hierarchical consistency of predictions~\cite{protect, park2024learning}, complemented by the leaf-level classification accuracy~\cite{vlm_img_bad, liu2024revisiting, he2025analyzing}, which serves as the upper bound of hierarchical consistency. The evaluation metrics are detailed below.

\noindent\textbf{Hierarchical Consistent Accuracy (HCA)} \cite{protect,park2024learning}. This metric is defined as 
\begin{align}
    \mathrm{HCA} = \frac{1}{N} \sum_{i=1}^{N} \prod_{j=1}^{L^i} \mathbbm{1}\left[f_\theta\left(x^i; \mathcal{Y}_{j}\right) = y^i_j\right], \label{eq:HCA}
\end{align}
Here, $N$ is the number of test samples, $L^i$ is the depth of the hierarchy for the $i$-th input $x^i$, and $\mathcal{Y}{j}$ denotes the label set at level~$j$. HCA computes the proportion of samples whose predicted paths exactly match the ground truth from root to leaf. It is therefore a stricter criterion than flat accuracy and serves as our primary evaluation metric for hierarchical classification.

\noindent\textbf{Point-Overlap Ratio (POR)}~\cite{por}.
It measures hierarchical performance beyond strict correctness, defined as:
\begin{equation}
\small
    \mathrm{POR} = \frac{1}{N} \sum_{i=1}^{N} \frac{\sum_{j=1}^{L_i} \mathbbm{1}\left[f_\theta\left(x_i; \mathcal{Y}_{j}\right) = y^i_j\right]}{L_i}.
\end{equation}
Unlike HCA, which requires an exact match along the entire path, POR allows partial correctness by averaging the proportion of correctly predicted nodes. This provides a fine-grained assessment of how well model outputs align with the target hierarchy.

\noindent\textbf{Strict Point-Overlap Ratio (S-POR).}
S-POR refines POR by rewarding only contiguous segments of correct predictions. For the $i$-th sample, we locate the longest run of consecutive correctly predicted nodes and normalize by the hierarchy depth $L_i$:

\begin{align}
\mathrm{S\text{-}POR}
&= \frac{1}{N}\sum_{i=1}^{N}\frac{1}{L_i}
  \max_{1\le a\le b\le L_i}
  \Bigl[(b-a+1)
  \notag\\[-3pt]
&\qquad\times
        \prod_{j=a}^{b}
        \mathbbm{1}\bigl[f_\theta(x_i;\mathcal{Y}_{j}) = y^i_j\bigr]\Bigr].
\end{align}
This stricter definition penalizes isolated correct predictions and emphasizes full-path consistency within the hierarchy.

\noindent\textbf{Top Overlap Ratio (TOR).}
Following \cite{protect}, TOR evaluates local hierarchical consistency by considering adjacent layer pairs as independent evaluation units:

\begin{equation}
\begin{aligned}
\small
\mathrm{TOR}
&= \frac{1}{N}\sum_{i=1}^{N}
    \frac{1}{L_i-1}\sum_{j=1}^{L_i-1}
    \mathbbm{1}\bigl[f_\theta(x_i;\mathcal{Y}_{j}) = y^i_j\bigr]
\\[-3pt]
&\quad\times
    \mathbbm{1}\bigl[f_\theta(x_i;\mathcal{Y}_{j+1}) = y^i_{j+1}\bigr].
\end{aligned}
\label{eq:tor}
\end{equation}
A TOR value of~1 indicates perfect pairwise consistency between consecutive layers, while lower scores reveal local violations of the hierarchical structure.

\subsection{Evaluation Setting}
\label{evaluation_setting}
For each setting, we treat large multimodal models (LMMs) as image classifiers $f_\theta$ and leverage language prompts to query predictions at different taxonomy levels. Following~\cite{tan2025vision}, we formulate HVR as a set of level-specific VQA tasks $(x^i, \mathcal{Y}_j)$, where $i=1,\ldots,N$ indexes images and $j=1,\ldots,L^i$ denotes taxonomy depth. To enable closed-set evaluation and avoid the ambiguity of open-set generation~\cite{vlm_img_bad}, each task is cast as a four-choice VQA problem:
\[
\begin{array}{l}
\text{{\textless image\textgreater Given the plant in the image, what is its taxonomic}}\\
\text{{classification at the \textless hierarchy\textgreater (e.g., kingdom) level?}}\\
\text{{A.\textless similar class\textgreater  \quad B.\textless ground truth\textgreater }}\\
\text{{C.\textless similar class\textgreater  \quad D.\textless similar class\textgreater }}\\
\text{{Answer with the option letter only.}}
\end{array}
\]

Although four-choice VQA is simpler than conventional hierarchical classification, whose label space grows exponentially with depth, we increase task difficulty by constructing semantically confusing distractors. Specifically, for each taxonomy level, we compute cosine similarity between the image and all incorrect labels using SigLIP~\cite{siglip}, and select the top three most similar labels as distractors. This ensures that all options lie at the same hierarchy level and exhibit strong semantic overlap with the ground truth.

\subsection{Implementation Details}
\label{implementation_datails}
All experiments are performed on 4×A6000 GPUs with a batch size of 1 per GPU and a two-step gradient accumulation. The models are all trained for 10 epochs. $\lambda_\mathrm{ent}$ and $\lambda_\mathrm{disp}$ are set to 0.1 and 0.01 by default. Curvature-aware scaling is applied to image embeddings with curvature $\kappa=0.05$.

\begin{table}[h]
  \centering
  \caption{Loss Coefficient $\lambda_\mathrm{ent}$ and $\lambda_\mathrm{disp}$.}
    \fontsize{8pt}{9pt}\selectfont
    \setlength{\tabcolsep}{8pt}
  \begin{tabular}{l|ccc}
    \toprule
      & $\lambda_\text{ent}$ = 0.05 & $\lambda_\text{ent}$ = 0.1 & $\lambda_\text{ent}$ = 0.2 \\
    \midrule
    $\lambda_\text{dis}$ = 0.005 & 17.40 & 16.22 & 17.18  \\
    $\lambda_\text{dis}$ = 0.01 & 16.55 & \sethlcolor{lightpink}\hl{18.07} & 16.77 \\
    $\lambda_\text{dis}$ = 0.02 & 16.45 & 17.55 & 16.25 \\
    \bottomrule
  \end{tabular}
  \vspace{.75em}

  \label{robust}
\end{table}

\subsection{Effects of Loss Coefficients $\lambda_\mathrm{ent}$ and $\lambda_\mathrm{disp}$}
\label{ablation_param}
We study the impact of the loss weights $\lambda_\mathrm{ent}$ and $\lambda_\mathrm{disp}$ in Table~\ref{tab:ablation_disp_variants}, Overall, all configurations studied improve upon the SFT baseline (HAC on base classes: 14.06), further suggesting that the regularizer is broadly effective. Model performance peaks at $\lambda_\mathrm{ent}=0.1$ and $\lambda_\mathrm{disp}=0.01$. Smaller values weaken the hierarchical and discriminative signals, while larger values can over-constrain the model and interfere with the primary language modeling objective.



\end{document}